\documentclass{article}
\usepackage[a4paper, margin=1in]{geometry}

\PassOptionsToPackage{numbers, compress}{natbib}

\usepackage{subcaption}

\usepackage{wrapfig}

\usepackage{enumitem}

\usepackage{placeins} 

\usepackage{soul} 

\setstcolor{red}

\usepackage[utf8]{inputenc}
\usepackage[T1]{fontenc} 
\usepackage{hyperref}  
\usepackage{url}     
\usepackage{booktabs}  
\usepackage{amsfonts}  
\usepackage{nicefrac} 
\usepackage{microtype}
\usepackage{xcolor}

\usepackage{graphicx}
\usepackage{amsmath}
\usepackage{amssymb}
\usepackage{natbib}
\usepackage{tikz}
\usetikzlibrary{fit, arrows, positioning, bending, shapes, shapes.geometric, arrows.meta}
\usepackage{amssymb}

\usepackage{mathtools}

\usepackage{tabularx}
\usepackage{booktabs}

\newcommand{\model}{Hierarchical Concept Memory Reasoner}
\newcommand{\acr}{H-CMR}

\usepackage{wrapfig}

\usepackage{pifont}
\newcommand{\cmark}{\textcolor{green}{\ding{51}}}
\newcommand{\xmark}{\textcolor{red}{\ding{55}}}
\newcommand{\hmark}{\textcolor{orange}{\boldsymbol{\sim}}}

\usepackage{amsthm}
\newtheorem{theorem}{Theorem}[section]

\usepackage{multirow}
\usepackage{bbm}

\usepackage{svg}
\usepackage{makecell} 

\usepackage{authblk}

\definecolor{green}{RGB}{100, 200, 100}

\definecolor{lightblue}{RGB}{173,216,230}

\title{Interpretable Hierarchical Concept Reasoning through Attention-Guided Graph Learning}

\author[1]{David Debot}
\author[2]{Pietro Barbiero}
\author[3]{Gabriele Dominici}
\author[1]{Giuseppe Marra}

\affil[1]{Department of Computer Science, KU Leuven}
\affil[2]{IBM Research, Zurich}
\affil[3]{Università della Svizzera Italiana, Lugano}  

\date{June~26,~2025}   

\begin{document}

\maketitle

\begin{abstract}
\noindent
Concept-Based Models (CBMs) are a class of deep learning models that provide interpretability by explaining predictions through high-level concepts. These models first predict concepts and then use them to perform a downstream task. However, current CBMs offer interpretability only for the final task prediction, while the concept predictions themselves are typically made via black-box neural networks. To address this limitation, we propose Hierarchical Concept Memory Reasoner (\acr{}), a new CBM that provides interpretability for both concept and task predictions. \acr{} models relationships between concepts using a learned directed acyclic graph, where edges represent logic rules that define concepts in terms of other concepts. During inference, \acr{} employs a neural attention mechanism to select a subset of these rules, which are then applied hierarchically to predict all concepts and the final task. Experimental results demonstrate that \acr{} matches state-of-the-art performance while enabling strong human interaction through concept and model interventions. The former can significantly improve accuracy at inference time, while the latter can enhance data efficiency during training when background knowledge is available.
\end{abstract}

\addtocontents{toc}{\protect\setcounter{tocdepth}{-1}}

\section{Introduction}

Concept-Based models (CBMs) have introduced a significant advancement in deep learning (DL) by making models explainable-by-design \cite{koh2020concept, alvarez2018towards, chen2020concept, EspinosaZarlenga2022cem, mahinpei2021promises, debot2024interpretable, barbiero2023interpretable, poeta2023concept, dominici2024causal, vandenhirtz2024stochastic, espinosa2023learning, havasi2022addressing}. These models integrate high-level, human-interpretable concepts directly into DL architectures, bridging the gap between black-box neural networks and transparent decision-making. One of the most well-known CBMs is the Concept Bottleneck Model (CBNM) \cite{koh2020concept}, which first maps an input (e.g.\ an image) to a set of human-understandable concepts (e.g.\ "pedestrian present," "danger in front") using a neural network and then maps these concepts to a downstream task (e.g.\ "press brakes") via a linear layer. The predicted concepts serve as an interpretable explanation for the final decision (e.g.\ "press brakes because there is a danger in front"). Considerable research has focused on ensuring CBMs achieve task accuracy comparable to black-box models. Some CBMs \cite{EspinosaZarlenga2022cem, mahinpei2021promises, debot2024interpretable, barbiero2023interpretable} are even known to be \textit{universal classifiers} \cite{hornik1989multilayer}; they match the expressivity of neural networks for classification tasks, regardless of the chosen set of concepts. 

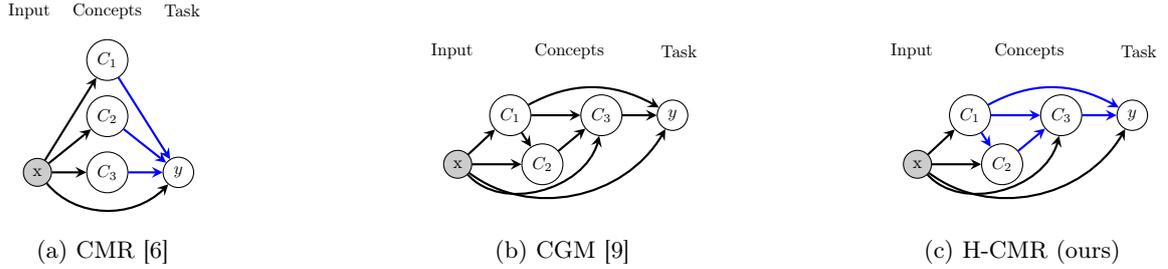
\begin{figure}
    \centering
    \begin{subfigure}{0.24\textwidth}
        \centering
        \begin{tikzpicture}[->, >=stealth, node distance=0.7cm]
            \begin{scope}[scale=0.65, transform shape]
                \node (X) [draw, circle, fill=black!20] {x};
                \node (C3) [draw, circle, right=of X] {$C_3$};
                \node (C2) [draw, circle, above=0.3cm of C3] {$C_2$};
                \node (C1) [draw, circle, above=0.3cm of C2] {$C_1$};
                \node (Y) [draw, circle, right=of C3] {$y$};

                \node (c) [above=0.3cm of C1] {Concepts};
                \node [right=0.2cm of c] {Task};
                \node [left=0.2cm of c] {Input};
    
                \draw[draw=black, thick] (X) -- (C1);
                \draw[draw=black, thick] (X) -- (C2);
                \draw[draw=black, thick] (X) -- (C3);
                \draw[color=blue, thick] (C1) -- (Y);
                \draw[color=blue, thick] (C2) -- (Y);
                \draw[color=blue, thick] (C3) -- (Y);
                \draw[draw=black, thick] (X) edge[bend right=50] (Y);
            \end{scope}
        \end{tikzpicture}
        \caption{CMR \cite{debot2024interpretable}}
        \label{fig:intro_diff2}
    \end{subfigure}
    \hfill
    \begin{subfigure}{0.24\textwidth}
        \centering
        \begin{tikzpicture}[->, >=stealth, node distance=0.7cm]
            \begin{scope}[scale=0.65, transform shape]
                \node (X) [draw, circle, fill=black!20] {x};
                \node (C2) [draw, circle, right=1cm of X] {$C_2$};
                \node (C1) [draw, circle, right=0.35cm of X, yshift=1cm] {$C_1$};
                \node (C3) [draw, circle, right=0.35cm of C2, yshift=1cm] {$C_3$};
                \node (Y) [draw, circle, right=of C3] {$y$};

                \node (c) [above=0.6cm of C1, xshift=1.2cm] {Concepts};
                \node [right=0.9cm of c] {Task};
                \node [left=1.0cm of c] {Input};
    
                \draw[draw=black, thick] (X) -- (C1);
                \draw[draw=black, thick] (X) -- (C2);
                \draw[draw=black, thick] (X) edge[bend right=65] (C3);
                \draw[draw=black, thick] (C1) -- (C3);
                \draw[draw=black, thick] (C1) -- (C2);
                \draw[draw=black, thick] (C2) -- (C3);
                \draw[draw=black, thick] (X) edge[bend right=50] (Y);
                \draw[draw=black, thick] (C1) edge[bend left=30] (Y);
                \draw[draw=black, thick] (C3) -- (Y);
            \end{scope}
        \end{tikzpicture}
        \caption{CGM \cite{dominici2024causal}}
        \label{fig:intro_diff3}
    \end{subfigure}
    \hfill
    \begin{subfigure}{0.24\textwidth}
        \centering
        \begin{tikzpicture}[->, >=stealth, node distance=0.7cm]
            \begin{scope}[scale=0.65, transform shape]
                \node (X) [draw, circle, fill=black!20] {x};
                \node (C2) [draw, circle, right=1cm of X] {$C_2$};
                \node (C1) [draw, circle, right=0.35cm of X, yshift=1cm] {$C_1$};
                \node (C3) [draw, circle, right=0.35cm of C2, yshift=1cm] {$C_3$};
                \node (Y) [draw, circle, right=of C3] {$y$};

                \node (c) [above=0.6cm of C1, xshift=1.2cm] {Concepts};
                \node [right=0.9cm of c] {Task};
                \node [left=1.0cm of c] {Input};
    
                \draw[draw=black, thick] (X) -- (C1);
                \draw[draw=black, thick] (X) -- (C2);
                \draw[draw=black, thick] (X) edge[bend right=65] (C3);
                \draw[color=blue, thick] (C1) -- (C3);
                \draw[color=blue, thick] (C1) -- (C2);
                \draw[color=blue, thick] (C2) -- (C3);
                \draw[draw=black, thick] (X) edge[bend right=50] (Y);
                \draw[color=blue, thick] (C1) edge[bend left=30] (Y);
                \draw[color=blue, thick] (C3) -- (Y);
            \end{scope}
        \end{tikzpicture}
        \caption{H-CMR (ours)}
        \label{fig:intro_diff4}
    \end{subfigure}
    
    \caption{Comparison of example CBMs. Blue and black edges are interpretable and black-box operations, respectively. (a) Some approaches model conditionally independent concepts with interpretable task inference. (b) Others learn a hierarchy of concepts with black-box task inference. (c) Only our approach learns a hierarchy with interpretable inference for both concepts and task.
    }
    \label{fig:intro_difference}
\end{figure}

\bigskip \noindent
While CBMs enhance interpretability at the task level, their concept predictions remain opaque, functioning as a black-box process. 
Most CBMs model the concepts as conditionally independent given the input (e.g.\ Figure \ref{fig:intro_diff2}), meaning any dependencies between them must be learned in an opaque way by the underlying neural network \cite{koh2020concept, alvarez2018towards, chen2020concept, EspinosaZarlenga2022cem, mahinpei2021promises, debot2024interpretable, barbiero2023interpretable}. In contrast, some existing approaches attempt to capture relationships between concepts in a structured way \cite{dominici2024causal} (e.g.\ Figure \ref{fig:intro_diff3}). However, the current approaches do not really provide interpretability: while they reveal which concept predictions influence others, they do not explain \textit{how} these influences occur. For instance, one can determine that 'pedestrian present' has some influence on 'danger in front,' but not the exact nature of this influence.

\bigskip \noindent
In this paper, we introduce \model{} (\acr{}) (Figure \ref{fig:intro_diff4}), the first CBM that is both a universal classifier and provides interpretability at both the concept and task levels. \acr{} learns a \textit{directed acyclic graph} (DAG) over concepts and tasks, which it leverages for inference. \acr{} employs \textit{neural rule generators} that produce symbolic logic rules defining how concepts and tasks should be predicted based on their parent concepts. These generators function as an attention mechanism between the input and a jointly learned memory of logic rules, selecting the most relevant rules for each individual prediction. Once the rules are selected for a specific input, the remaining inference is entirely interpretable for both concepts and tasks, as it follows straightforward logical reasoning. This means the human can inspect how parent concepts exactly contribute to their children. By combining graph learning, rule learning, and neural attention, \acr{} provides a more transparent and structured framework for both concept predictions and downstream decision-making.

\bigskip \noindent
Our experiments demonstrate that \acr{} achieves state-of-the-art accuracy while uniquely offering interpretability for both concept and task prediction. This interpretability enables strong human-AI interaction, particularly through interventions. First, humans can do \textit{concept interventions} at inference time, correcting mispredicted concepts. Unlike in typical CBMs, which model concepts as conditionally independent, these interventions are highly effective: correcting one concept can propagate to its dependent concepts, potentially cascading through multiple levels. Second, humans can do \textit{model interventions} at training time, modifying the graph and rules that are being learned, allowing the human to shape (parts of) the model. We show that this enables the integration of prior domain knowledge, improving data efficiency and allowing \acr{} to perform in low-data regimes.

\bigskip \noindent
We want to stress that \acr{} does not attempt to learn the causal dependencies between concepts \textit{in the data}.
Instead, \acr{} reveals its internal reasoning process: which concepts does the model use for predicting other concepts (also known as \textit{causal transparency} \cite{dominici2024causal}), and how (\textit{interpretability}).

\section{Model}

We introduce \model{} (\acr{}), the first CBM that is a universal binary classifier providing interpretability for both concept and task predictions. 
We considered three main desiderata when designing \acr{} (for more details, see Section \ref{sec:dimensions}): 
\begin{itemize}
    \item \textbf{interpretability}, the ability for humans to understand how concepts and tasks are predicted using each other; 
    \item \textbf{intervenability}, the ability for humans to meaningfully interact with the model;
    \item \textbf{expressivity}, a requirement for the model to be able to achieve similar levels of accuracy as black-box models irrespective of the employed set of concepts.
\end{itemize} 
For simplicity, we only consider concepts in the remaining sections, omitting the task. The task can be treated similar as the concepts, or separately like in most CBMs (see Appendix \ref{app:props}).

\subsection{High-level overview}

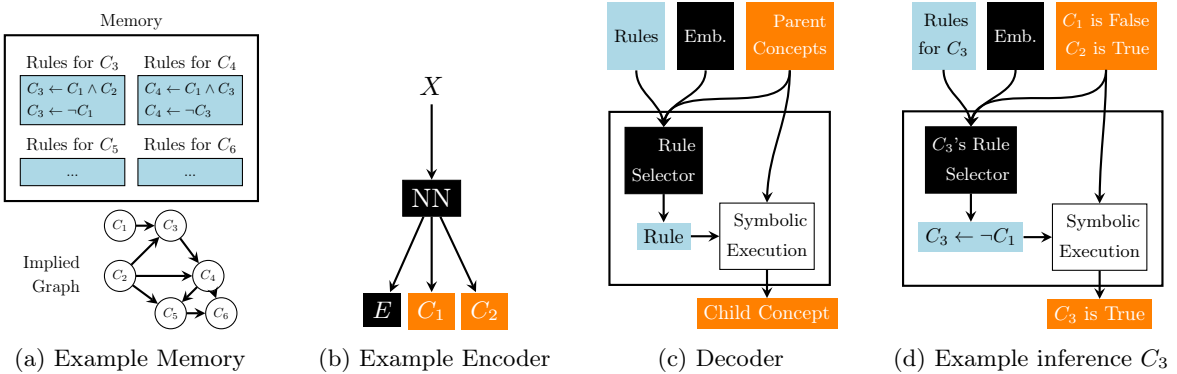
\begin{figure}
    \centering
    \begin{subfigure}[b]{0.27\textwidth}
        \centering
        \begin{tikzpicture}[->, >=stealth, node distance=0.2cm]
            \begin{scope}[scale=0.55, transform shape]

                \node (Memory1) [draw, fill=lightblue, rectangle, minimum width=2.5cm] {
                    $\begin{aligned} 
                        C_3 &\leftarrow C_1 \land C_2 \\ 
                        C_3 &\leftarrow \neg C_1
                    \end{aligned}$
                    };
                \node (c3rules) [font=\large, above=0.0cm of Memory1] {Rules for $C_3$};

                \node (Memory2) [draw, fill=lightblue, rectangle, minimum width=2.5cm, right=0.3cm of Memory1] {
                    $\begin{aligned} 
                        C_4 &\leftarrow C_1 \land C_3 \\ 
                        C_4 &\leftarrow \neg C_3
                    \end{aligned}$
                    };
                \node [font=\large, above=0.0cm of Memory2] {Rules for $C_4$};

                \node (Memory3) [draw, fill=lightblue, rectangle, minimum width=2.5cm, below=0.8cm of Memory1] {
                    $\begin{aligned} 
                        ...
                    \end{aligned}$
                    };
                \node [font=\large, above=0.0cm of Memory3] {Rules for $C_5$};

                \node (Memory4) [draw, fill=lightblue, rectangle, minimum width=2.5cm, below=0.8cm of Memory2] {
                    $\begin{aligned} 
                        ...
                    \end{aligned}$
                    };
                \node [font=\large, above=0.0cm of Memory4] {Rules for $C_6$};

                \node (gr) [font=\large, below=1.5cm of Memory3, xshift=-0.5cm] {
                    $\begin{aligned}
                        \text{Implied}  \\
                        \text{Graph}
                    \end{aligned}$
                };

                \node[draw, thick, fit=(Memory1)(Memory2)(Memory3)(Memory4)(c3rules), inner sep=0.2cm, label=above:{\large Memory}] {};

                \begin{scope}[scale=0.9, transform shape]
                    \node (C2) [draw, circle, right=0.5cm of gr] {$C_2$};
                    \node (C1) [draw, circle, above=0.5cm of C2] {$C_1$};
                    \node (C3) [draw, circle, right=0.5cm of C1] {$C_3$};
                    \node (C4) [draw, circle, right=1.5cm of C2] {$C_4$};
                    \node (C5) [draw, circle, below=1.5cm of C3] {$C_5$};
                    \node (C6) [draw, circle, right=0.5cm of C5] {$C_6$};
        
                    \draw[draw=black, thick] (C1) -- (C3);
                    \draw[draw=black, thick] (C2) -- (C3);
                    \draw[draw=black, thick] (C3) -- (C4);
                    \draw[draw=black, thick] (C2) -- (C4);
                    \draw[draw=black, thick] (C2) -- (C5);
                    \draw[draw=black, thick] (C4) -- (C5);
                    \draw[draw=black, thick] (C4) -- (C6);
                    \draw[draw=black, thick] (C5) -- (C6);
                \end{scope}
                
            \end{scope}
        \end{tikzpicture}
        \caption{Example Memory}
        \label{fig:overview_memory}
    \end{subfigure}
    \hfill
    \begin{subfigure}[b]{0.20\textwidth}
        \centering
        \begin{tikzpicture}[->, >=stealth, node distance=0.2cm]
            \begin{scope}[scale=1, transform shape]
                \node (x) {$X$};
                \node (nn) [draw, fill=black, text=white, below=1cm of x] {NN};
                \node (e) [fill=black, text=white, below=1cm of nn, xshift=-0.65cm] {\small $E$};
                \node (c1) [fill=orange, text=white, below=1cm of nn] {\small $C_1$};
                \node (c2) [fill=orange, text=white, below=1cm of nn, xshift=0.7cm] {\small $C_2$};

                \draw[thick] (x) -- (nn);
                \draw[thick] (nn) -- (c1);
                \draw[thick] (nn) -- (c2);
                \draw[thick] (nn) -- (e);
            \end{scope}
        \end{tikzpicture}
        \caption{Example Encoder}
    \end{subfigure}
    \hfill
    \begin{subfigure}[b]{0.24\textwidth}
        \centering
        \begin{tikzpicture}[->, >=stealth, node distance=0.2cm]
            \begin{scope}[scale=0.75, transform shape]
                \node (parents) [fill=orange, text=white, minimum height=1.2cm] {\small $\begin{aligned}
                        \text{Parent} \\
                        \text{Concepts}
                    \end{aligned}$};
                \node (e) [fill=black, text=white, left=0.2cm of parents, minimum height=1.2cm] {\small Emb.\ };
                \node (rules) [fill=lightblue, left=0.2cm of e, minimum height=1.2cm] {\small Rules};
                \node (selector) [draw, fill=black, text=white, below=1.0cm of rules, xshift=0.48cm] {
                    $\begin{aligned}
                        \small \text{Rule} \\
                        \small \text{Selector}
                    \end{aligned}$
                };
                \node (rule) [fill=lightblue, text=black, below=0.5cm of selector] {Rule};
                \node (exec) [draw, right=0.5cm of rule] {
                $\begin{aligned}
                    \small \text{Symbolic} \\
                    \small \text{Execution}
                \end{aligned}$
                };
                \node (concept) [fill=orange, text=white, below=0.5cm of exec] {Child Concept};

                \node[draw, thick, fit=(selector)(rule)(exec), inner sep=0.2cm] {};

                \draw[thick] (rules) edge[in=90, out=-90] (selector);
                \draw[thick] (e) edge[in=90, out=-90] (selector);
                \draw[thick] (parents) edge[in=90, out=-90] (selector);
                \draw[thick] (selector) -- (rule);
                \draw[thick] (parents) edge[in=90, out=-90] (exec);
                \draw[thick] (rule) -- (exec);
                \draw[thick] (exec) -- (concept);
            \end{scope}
        \end{tikzpicture}
        \caption{Decoder}
    \end{subfigure}
    \hfill
    \begin{subfigure}[b]{0.24\textwidth}
        \centering
        \begin{tikzpicture}[->, >=stealth, node distance=0.2cm]
            \begin{scope}[scale=0.75, transform shape]
                \node (parents) [fill=orange, text=white, minimum height=1.2cm] {\small 
                    $\begin{aligned}
                        \text{$C_1$ is False} \\
                        \text{$C_2$ is True}
                    \end{aligned}$};
                \node (e) [fill=black, text=white, left=0.2cm of parents, minimum height=1.2cm] {\small Emb.\ };
                \node (rules) [fill=lightblue, left=0.2cm of e, minimum height=1.2cm] {\small 
                    $\begin{aligned}
                        \text{Rules} \\
                        \text{for $C_3$}
                    \end{aligned}$};
                \node (selector) [draw, fill=black, text=white, below=1.0cm of rules, xshift=0.48cm] {
                    $\begin{aligned}
                        \small \text{$C_3$'s Rule} \\
                        \small \text{Selector}
                    \end{aligned}$
                };
                \node (rule) [fill=lightblue, below=0.5cm of selector] {$C_3 \leftarrow \neg C_1$};
                \node (exec) [draw, right=0.5cm of rule] {
                $\begin{aligned}
                    \small \text{Symbolic} \\
                    \small \text{Execution}
                \end{aligned}$
                };
                \node (concept) [fill=orange, text=white, below=0.5cm of exec] {$C_3$ is True};

                \node[draw, thick, fit=(selector)(rule)(exec), inner sep=0.2cm] {};

                \draw[thick] (rules) edge[in=90, out=-90] (selector);
                \draw[thick] (e) edge[in=90, out=-90] (selector);
                \draw[thick] (parents) edge[in=90, out=-90] (selector);
                \draw[thick] (selector) -- (rule);
                \draw[thick] (parents) edge[in=90, out=-90] (exec);
                \draw[thick] (rule) -- (exec);
                \draw[thick] (exec) -- (concept);
            \end{scope}
        \end{tikzpicture}
        \caption{Example inference $C_3$}
    \end{subfigure}

    \caption{High-level overview of the different components of \acr{}. (a) The memory is compartmentalized per concept and implies a DAG over the concepts. (b) The encoder predicts source concepts and an embedding. (c) The decoder infers each non-source concept from its parent concepts, the embedding and its rules, by selecting a rule (using a neural network) and then symbolically executing that rule using the parent concepts. (d) Example of the decoder predicting $C_3$.}
    \label{fig:overview_blocks}
\end{figure}

From a high-level perspective, \acr{} consists of three main components (Figure \ref{fig:overview_blocks}): a \textit{rule memory}, an \textit{encoder}, and a \textit{decoder}. The encoder predicts a small number of concepts and an embedding, and the decoder selects from the memory a set of logic rules to hierarchically predict all other concepts.

\bigskip \noindent
\textbf{Memory.} The memory is compartmentalized per concept. For each concept, it stores a set of (learned) logic rules that define that concept in terms of other concepts, e.g. $C_3 \leftarrow C_1 \land C_2$, or $C_3 \leftarrow \neg C_1$.  This memory implicitly defines a \textit{directed acyclic graph} (DAG) over the concepts: if a concept $C_i$ appears in at least one rule defining another concept $C_j$, then $C_i$ is a parent of $C_j$. For some concepts, all associated rules will be "empty" (e.g.\ $C_i \leftarrow .$), meaning they have no parents. These are the \textit{source concepts} of the DAG, which are predicted directly by the encoder rather than inferred via rules.

\bigskip \noindent
\textbf{Encoder.} The encoder is a neural network which maps the input to the source concepts and a latent embedding. Thus, source concepts are directly predicted from the input in the same black-box fashion done by standard CBMs.\footnote{While standard CBMs do this for all concepts, \acr{} only does this for source concepts.} The latent embedding captures additional contextual information from the input that may not be captured by the concepts. This preserves the concept-prediction expressivity of other CBMs, and the task-prediction expressivity of black-box neural networks (see Section \ref{sec:dimensions}).

\bigskip \noindent
\textbf{Decoder.} The decoder is used to hierarchically perform inference over the concept DAG. At each step, it predicts a concept using its parent concepts, the latent embedding, and its rules in the memory. It leverages a neural attention mechanism to select the most relevant rule for the current prediction, based on the parent concepts and the latent embedding. This rule is then symbolically executed on the parent concepts to produce the concept prediction.

\bigskip \noindent
This approach is designed to handle settings where (i) no graph over concepts is available and must therefore be learned from data, (ii) rules defining concepts in terms of others are unknown and must be learned, and (iii) the available concepts alone are \textit{insufficient} for perfect prediction, requiring additional contextual information, here exploited through the rule selection and the latent embedding. While the learned rules may be noisy in settings where concepts are insufficient (i.e.\ directly applying all rules may not lead to correct predictions in every case), this is not a problem due to the selection mechanism, as only the \textit{selected rule} is required to yield the correct prediction.

\subsection{Parametrization}

In this section, we go into more detail on how the individual components are parametrized. We refer to Appendix \ref{app:pgm} for \acr{}'s probabilistic graphical model, and Appendix \ref{app:exp_details} for more details regarding the neural network architectures.
We explain how these components are used to do inference in Section \ref{sec:inference}.

\subsubsection{Encoder}

The encoder directly predicts each source concept $C_i$ and the embedding $E$ from the input $x$:
\begin{equation}  \label{eq:root_concept_node}
    p(C_i = 1 \mid \hat{x}) = f_i(\hat{x}), \quad \hat{e} = g(\hat{x})
\end{equation}
where each $f_i$ and $g$ are neural networks, with the former parametrizing Bernoulli distributions.\footnote{We abbevriate the notation for assignments to random variables, e.g.\ $\hat{x}$ means $X=\hat{x}$.} This deterministic modelling of the embedding $E$ corresponds to a delta distribution.

\subsubsection{Decoder}

The decoder infers non-source concepts from their parent concepts, the latent embedding and the rules in the memory, and operates in two steps. First, the parent concepts and the embedding are used to select a logic rule from the set of rules for that concept. Second, this rule is evaluated on the parent concept nodes' values to produce an interpretable prediction. More details on this memory and the representation and evaluation of rules are given in Section \ref{sec:memory}. 

\bigskip \noindent
The selection of a rule for a concept $C_i$ is modelled as a categorical random variable $S_i$ with one value per rule for $C_i$. For instance, if there are three rules for $C_1$ and the predicted categorical distribution for $S_1$ is $(0.8, 0.2, 0.0)$, then this means that the first rule in the memory is selected with 80\% probability, the second rule with 20\%, and the third with 0\%. The logits of this distribution are parametrized by a neural network that takes the parent concepts and latent embedding as input. For each non-source concept $C_i$, the concept prediction is:
\begin{align}  \label{eq:non_root_concept_node}
    p(C_i = 1 \mid \hat{e}, \hat{c}_{parents(i)}, \hat{r}_i) = \sum_{k=1}^{n_{R}} \,\,\,\, \underbrace{p(S_i=k\; \mid \hat{e}, \hat{c}_{parents(i)})}_{\mathclap{\substack{\text{neural selection of rule} \, k \\ \text{using parent concepts + emb.}}}} \,\,\,\, \cdot \,\,\,\, \underbrace{l(\hat{c}_{parents(i)}, \hat{r}_{i,k})}_{\mathclap{\substack{\text{evaluation of concept $i$'s rule} \, k \\ \text{using parent concepts}}}}
\end{align}
with $\hat{e}$ the latent embedding, $n_{R}$ the number of rules for each concept, $\hat{r}_i$ the set of rules for this concept, $S_i$ the predicted categorical distribution over these rules, and $l(\hat{c}, \hat{r}_{i,k})$ the symbolic execution of rule $k$ of concept $i$ using concepts $\hat{c}$ (see Section \ref{sec:memory}). Intuitively, all the learned rules for $C_i$ contribute to its prediction, each weighted by the rule probability according to the neural selection.

\subsubsection{Memory, rule representation and rule evaluation} \label{sec:memory}

For each concept, \acr{} learns $n_R$ rules in its memory, with $n_R$ a hyperparameter. The memory and the representation of rules resemble the approach of \cite{debot2024interpretable}.\footnote{Note that their rules define tasks in terms of concepts. Ours also define concepts in terms of each other.} For each concept $C_i$, the memory contains $n_R$ embeddings, each acting as a latent representation of a rule. These embeddings are decoded using a neural network into symbolic representations of logic rules, enabling symbolic inference. We consider rule bodies that are conjunctions of concepts or their negations, e.g.\ $C_3 \leftarrow C_0 \land \neg C_1$ (read "if $C_0$ is true and $C_1$ is false, then $C_3$ is true"). 

\bigskip \noindent
A rule is represented as an assignment to a categorical variable over all possible rules. Explicitly defining this distribution would be intractable, as there are an exponential number of possible rules. Instead, we factorize this variable into $n_C$ independent categorical variables $R_i$, each with 3 possible values corresponding to the \textit{role} of a concept in the rule. For instance, in $C_3 \leftarrow C_0 \land \neg C_1$, we say $C_0$ plays a \textit{positive} ($R_0=P$) role, $C_1$ plays a \textit{negative} ($R_1=N$) role, and $C_2$ is \textit{irrelevant} ($R_2=I$).

\bigskip \noindent
Evaluating a rule on concept predictions follows the standard semantics of the logical connectives. Using our representation of a rule, this becomes: 
\begin{equation}  \label{eq:logic_eval}
l(\hat{c}, \hat{r}_{i,k}) = \prod_{j=1}^{n_C} (\mathbbm{1}[\hat{r}_{i,k,j} = P] \cdot \mathbbm{1}[\hat{c}_j=1] + \mathbbm{1}[\hat{r}_{i,k,j} = N]  \cdot \mathbbm{1}[\hat{c}_j=0 ])
\end{equation}
where $n_C$ is the number of concepts, $l(\cdot)$ is the logical evaluation of a given rule using the given concepts $\hat{c}$, and $\hat{r}_{i,k,j}$ is the role of concept \textit{j} in that rule (positive (P), negative (N) or irrelevant (I)).

\begin{figure}
    \centering
    
    \hfill
    \begin{subfigure}{0.3\textwidth}
        \centering
        \begin{tikzpicture}[->, >=stealth, node distance=0.3cm]
            \begin{scope}[scale=0.7, transform shape]
            \node (C0) [draw, circle] {$C_0$};
            \node (C1) [draw, circle, below=of C0)] {$C_1$};
            \node (C3) [draw, circle, right=of C1] {$C_3$};
            \node (C2) [draw, circle, right=of C0] {$C_2$};

            \draw[draw=black, thick] (C0) -- (C1);
            \draw[draw=black, thick] (C0) -- (C2);
            \draw[draw=black, thick] (C0) -- (C3);
            \draw[draw=black, thick] (C1) -- (C2);
            \draw[draw=black, thick] (C1) -- (C3);
            \draw[draw=black, thick] (C2) -- (C3);
            \end{scope}
        \end{tikzpicture}
        \caption{All possible edges allowed by the node priorities ($O$).}
    \end{subfigure}
    \hspace{10pt}
    \begin{subfigure}{0.3\textwidth}
        \centering
        \begin{tikzpicture}[->, >=stealth, node distance=0.3cm]
            \begin{scope}[scale=0.7, transform shape]
            \node (C0) [draw, circle] {$C_0$};
            \node (C1) [draw, circle, below=of C0)] {$C_1$};
            \node (C3) [draw, circle, right=of C1] {$C_3$};
            \node (C2) [draw, circle, right=of C0] {$C_2$};

            \draw[draw=black, thick] (C0) -- (C3);
            \draw[draw=black, thick] (C0) -- (C2);
            \draw[draw=black, thick] (C1) -- (C0);
            \draw[draw=black, thick] (C1) -- (C3);
            \draw[draw=black, thick] (C2) -- (C3);
            \draw[draw=black, thick] (C2) edge[bend right=45] (C0);
            \end{scope}
        \end{tikzpicture}
        \caption{Example graph imposed by unadjusted rules ($R'$).}
    \end{subfigure}
    \hspace{10pt}
    \begin{subfigure}{0.3\textwidth}
        \centering
        \begin{tikzpicture}[->, >=stealth, node distance=0.3cm]
            \begin{scope}[scale=0.7, transform shape]
            \node (C0) [draw, circle] {$C_0$};
            \node (C1) [draw, circle, below=of C0)] {$C_1$};
            \node (C3) [draw, circle, right=of C1] {$C_3$};
            \node (C2) [draw, circle, right=of C0] {$C_2$};

            \draw[draw=black, thick] (C0) -- (C3);
            \draw[draw=black, thick] (C0) -- (C2);
            \draw[draw=black, thick] (C1) -- (C3);
            \draw[draw=black, thick] (C2) -- (C3);
            \end{scope}
        \end{tikzpicture}
        \caption{Example DAG imposed by adjusted rules ($R$).}
    \end{subfigure}
    
    \caption{Example of \acr{}'s learned DAG over concepts as defined by its learned rules. The learned node priority vector $O$ (i.c.\ $O_0 < O_1 < O_2 < O_3$) enforces a topological ordering of the nodes, guaranteeing that the learned graph is a DAG.}
    \label{fig:dag_adjusted_rules}
\end{figure}
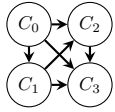
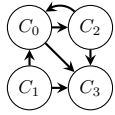
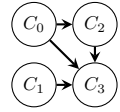

\bigskip \noindent
Decoding each rule embedding into this symbolic representation (i.e.\ assignments to $n_C$ categorical variables $R_i$) happens in two steps, and is different from \cite{debot2024interpretable}. First, a neural network maps each rule embedding to the logits for $n_C$ categorical distributions $R'_i$.
Then, to ensure that the rules form a DAG, we must prevent cyclic dependencies. For instance, we should not learn conflicting rules such as $C_1 \leftarrow C_0$ and $C_0 \leftarrow C_1$, or $C_1 \leftarrow C_1$. To enforce this constraint, we draw inspiration from \cite{massidda2023constraint}, defining a learnable node priority vector which establishes a topological ordering over concepts: higher-priority concepts are not allowed to appear in the rules of lower-priority concepts, thereby preventing cycles. We achieve this by using the node priorities to modify the categorical distributions $R'_i$, obtaining the to-be-used distributions $R_i$. Specifically, we make any rule that violates the ordering impossible, ensuring its probability is zero. Let $O_i$ be the node priority of concept $i$, then:
\begin{align}  \label{eq:true_roles}
    \forall r \in \{P, N\}: p(R_{i,k,j}=r) &= \mathbbm{1}[O_j>O_i] \cdot p(R'_{i,k,j}=r) \\
    p(R_{i,k,j}=I) &= \mathbbm{1}[O_j>O_i] \cdot p(R'_{i,k,j}=I) + \mathbbm{1}[O_j \leq O_i]
\end{align}
where $\mathbbm{1}[\cdot]$ is the indicator function, and $p(R_{i,k,j})$ is the categorical distribution of the role of concept $j$ in rule $k$ for concept $i$ corrected with the node priorities $O$, which are modelled as delta distributions. Figure \ref{fig:dag_adjusted_rules} gives a graphical example. The employed rules in the memory are assignments to these random variables, which are used in Equation \ref{eq:logic_eval}. During training, these assignments are obtained by sampling from this distribution (see Section \ref{sec:learning}). During inference, we simply take the most likely roles (see Section \ref{sec:inference}). 

\section{Inference}  \label{sec:inference}

For the derivation of the equations below from \acr{}'s probabilistic graphical model, we refer to Appendix \ref{app:pgm}. Computing the exact likelihood of a concept corresponds to:
\begin{align}  \label{eq:inference_complete}
    p(C_i|\hat{x}) = \sum_{\hat{c}_{parents(i)}} & \mathbbm{1}[Source_i=1] \cdot p(C'_i \mid \hat{x}) + \mathbbm{1}[Source_i=0] \cdot p(C_i'' \mid \hat{x}, \hat{c}_{parents(i)}, \hat{r}_i)
\end{align}
where $\hat{r}_i$ is given by Equation \ref{eq:roles_argmaxed}, $p(C'_i \mid \cdot)$ by Equation \ref{eq:root_concept_node} and $p(\hat{C_i''} \mid \cdot)$ by Equation \ref{eq:non_root_concept_node}. Concepts are source concepts if all of their rules are empty, i.e.\ for all their rules, each concept is irrelevant: 
$Source_i  = \prod_{k=1}^{n_R} \prod_{j=1}^{n_C} \mathbbm{1}[\hat{r}_{i,k,j}=I]$.
The sum goes over all possible assignments to the parent concepts.
As this would make inference intractable, we instead take an approximation of the Maximum A Posteriori estimate over the concepts by thresholding each individual concept prediction at 50\%:\footnote{An alternative is to sample the concepts instead.}
\begin{align} \label{eq:inference}
    p(C_i|\hat{x}) = \; & \mathbbm{1}[Source_i=1] \cdot p(C'_i \mid \hat{x}) + \mathbbm{1}[Source_i=0] \cdot p(C_i'' \mid \hat{x}, \hat{c}_{parents(i)}, \hat{r}_i)
\end{align}
with $\hat{c}_{parents(i)} = \{ \mathbbm{1}[p(C_j=1 \mid \hat{x}) > 0.5] \mid \mathbbm{1}[Parent_{ij}=1] \}$. 
Note that this thresholding is also beneficial for avoiding the problem of concept leakage in CBMs, which harms interpretability \cite{marconato2022glancenets}.
As mentioned earlier, a concept is another concept's parent if it appears in at least one of its rules: $Parent_{ij}=1-\prod_{k=1}^{n_R} \mathbbm{1}[\hat{r}_{i,k,j}=I]$ (i.e.\ in not all rules for $C_i$, the role of $C_j$ is irrelevant). 
Finally, the roles $\hat{r}_i$, which represent logic rules, are the most likely roles:
\begin{align}  \label{eq:roles_argmaxed}
    \hat{r}_{i,k,j} &= \underset{r \in \{P, N, I\}}{\text{arg max}} \, p(R_{i,k,j}=r)
\end{align}

\section{Learning problem}  \label{sec:learning}

During learning, \acr{} is optimized jointly: the encoder (neural network), the decoder (rule selector neural networks) and the memory (node priority vector, rule embeddings and rule decoding neural networks). The training objective follows a standard objective for CBMs, maximizing the likelihood of the concepts. For the derivation of this likelihood and the other equations below using \acr{}'s probabilistic graphical model, we refer to Appendix \ref{app:pgm}. Because the concepts are observed during training, this likelihood becomes:
\begin{equation}  \label{eq:factorized_objective}
    \max_{\Omega} \sum_{(\hat{x},\hat{c}) \in \mathcal{D}} \sum_{i=1}^{n_C} \log p(\hat{c}_i \mid \hat{c}, \hat{x})
\end{equation}
where we write $\hat{c}_{parents(i)}$ as $\hat{c}$ to keep notation simple. Each individual probability is computed using Equation \ref{eq:root_concept_node} for source concepts and Equation \ref{eq:non_root_concept_node} for other concepts:
\begin{align}  \label{eq:learning}
    p(\hat{c}_i|\hat{c}, \hat{x}) & = \mathbb{E}_{\hat{r} \sim p(R)} \left[  p(\hat{c}'_i \mid \hat{x}) \cdot p(Source_i \mid \hat{r}) + p(\hat{c}''_i \mid \hat{c}, \hat{e}, \hat{r}) \cdot p(\neg Source_i \mid \hat{r})   \right] \\
    \text{where} & \quad \quad p(Source_i \mid \hat{r}) = \prod_{j=1}^{n_C} \prod_{k=1}^{n_R} \mathbbm{1}[R_{i,k,j}=I]
\end{align}
with $p(C_i'\mid \hat{x})$ and $\hat{e}$ corresponding to Equation \ref{eq:root_concept_node}, $p(C_i'' \mid \hat{c}, \hat{e}, \hat{r})$ to Equation \ref{eq:non_root_concept_node}, and $p(R)$ to Equation \ref{eq:true_roles}.\footnote{We use straight-through estimation for the thresholding operator in Equation \ref{eq:true_roles} and the sampling of $R$.} Note that the designation of source concepts and parent concepts may change during training, as the roles $R$ change.
Additionally, to promote learning rules that are \textit{prototypical} of the seen concepts, we employ a form of regularization akin to \cite{debot2024interpretable} (see Appendix \ref{app:exp_details}).

\bigskip \noindent
\textbf{Scalability.} Computing the above likelihood scales $O(n_R \cdot n_C^2)$ at training time with $n_C$ the number of concepts and $n_R$ the number of rules per concept. At inference time, this becomes the worst-case complexity, depending on the structure of the learned graph.

\section{Expressivity, interpretability and intervenability} \label{sec:dimensions}

\subsection{Expressivity}

\acr{} functions as a universal binary classifier for both concept and task prediction. This means it has the same expressivity of a neural network classifier, regardless of the employed concepts. In practice, this translates to high accuracy across a wide range of applications.
\begin{theorem}
    \label{th:expressivity}
    \acr{} is a universal binary classifier \cite{hornik1989multilayer} if $n_R \geq 2$, with $n_R$ the number of rules for each concept and task.
\end{theorem}
\noindent
Furthermore, \acr{}'s parametrization guarantees that the learned graph over the concepts forms a DAG, and is expressive enough to represent any possible DAG. Let $\Theta$ be the set of all possible parameter values \acr{} can take. For a specific parameter assignment $\theta \in \Theta$, let $\mathcal{G}_\theta$ represent the corresponding concept graph.
\begin{theorem}
    \label{th:expressivity_graph}
    Let $\mathcal{DAG}$ denote the set of all directed acylic graphs (DAGs). Let $\mathcal{H} \coloneq \{ \mathcal{G}_\theta \mid \theta \in \Theta \}$ be the set of graphs over concepts representable by \acr{}. Then:
    \[ \mathcal{H} = \mathcal{DAG} \]
    That is,
    \[ \forall \, G \in \mathcal{DAG}, \exists \, \theta \in \Theta: \mathcal{G}_\theta=G, \quad \forall \theta \in \Theta: \mathcal{G}_\theta \, \, \text{is a DAG} \]
\end{theorem}
\noindent
For the proofs, we refer to Appendix \ref{app:proofs}.

\subsection{Interpretability}

In sharp contrast to other CBMs, \acr{} offers interpretability not only for task prediction but also for concept prediction. Most CBMs model the concepts as conditionally independent given the input, leading to their direct prediction through an uninterpretable black-box mechanism. In \acr{}, interpretability is achieved by representing concepts as the logical evaluation of a neurally selected rule. \acr{} provides two distinct forms of interpretability: \textit{local} and \textit{global}.

\bigskip \noindent
\acr{} provides local interpretability by making the logic rules used for predicting both concepts and tasks explicitly transparent to the human for a given input. Once these rules are selected for a given input instance, the remaining computation is inherently interpretable, as it consists of logical inference over the structure of the graph using these rules.

\bigskip \noindent
\acr{} enables a form of global interpretability, as all possible rules applicable for obtaining each concept and task prediction are stored transparently in the memory. First, this allows for human inspection of the rules, and even formal verification against predefined constraints using automated tools (see Appendix \ref{app:props}). Second, this allows for \textit{model interventions} (see Section \ref{sec:intervenability}).

\subsection{Intervenability} \label{sec:intervenability}

\textbf{Concept interventions.} Concept interventions are test-time operations in which some concept predictions are replaced with their ground truth values. These interventions simulate interactions with human experts at decision time and are considered a crucial feature of CBMs. Ideally, interventions should have the greatest possible impact on the model's predictions. 
\acr{} does not model concepts as conditionally independent given the input, which provides a significant advantage when performing interventions over most CBMs: \acr{} allows interventions on parent concepts to propagate their effects to child concepts, which in turn can influence further downstream concepts. As a result, \acr{} demonstrates greater responsiveness to interventions compared to CBMs where concepts are conditionally independent of one another. 
Specifically, an intervention on a concept can affect child concept predictions in two  ways. First, it can modify the rule selection process for the child concept, because the intervened concept is an input to the child concept's rule selection. Second, it can alter the evaluation of the selected logic rule, as the concept may be used in evaluating that rule. 

\bigskip \noindent
\textbf{Model interventions.} In addition to concept interventions at test time, \acr{} allows for model interventions at training time, influencing the graph and rules that are learned. 
A human expert's knowledge can be incorporated by manually adding new rules to the memory, and learned rules can be inspected, modified, or replaced as needed. Moreover, the human can forbid concepts from being parents of other concepts, or enforce specific structures on the graph (e.g.\ choose which concepts should be sources or sinks). For details on how this can be done, we refer to Appendix \ref{app:props}.

\section{Experiments}

In our experiments, we consider the following research questions: \textbf{(Accuracy)} Does \acr{} attain similar concept accuracy as existing CBMs? Does \acr{} achieve high task accuracy irrespective of the concept set? \textbf{(Explainability and intervenability)} Does \acr{} learn meaningful rules? Are concept interventions effective? Can model interventions be used to improve data efficiency?

\subsection{Experimental setting}

We only list essential information for understanding the experiments. Details can be found in Appendix \ref{app:exp_details}. We focus on concept prediction as this is what distinguishes \acr{} the most, omitting the task for most experiments and comparing with state-of-the-art concept predictors. In one experiment we still support our claim that \acr{} can achieve high task accuracy irrespective of the concept set.

\bigskip \noindent
\textbf{Data and tasks.}  
We use four datasets to evaluate our approach: CUB \citep{cub}, a dataset for bird classification;
MNIST-Addition \cite{manhaeve2018deepproblog}, a dataset based on MNIST \cite{mnistLecun}; CIFAR10 \cite{krizhevsky2009learning}, a widely used dataset in machine learning; and a synthetic dataset based on MNIST with difficult concept prediction and strong dependencies between concepts. 
All datasets except CIFAR10 provide full concept annotations. For CIFAR10, we use the same technique as \cite{oikarinen2023labelfree} to extract concept annotations from a vision-language model, showing our approach also works on non-concept-based datasets.\footnote{Note that our approach can be applied to multilabel classification dataset \textit{without any concepts}, in which case \acr{} learns a graph over the different tasks (instead of over concepts and tasks).}

\bigskip \noindent
\textbf{Evaluation.} 
We measure classification performance using accuracy. For intervenability, we report the effect on accuracy of intervening on concepts (see Appendix \ref{app:more_results} for the used intervention strategies and ablation studies). Metrics are reported using the mean and standard deviation over 3 seeded runs.

\bigskip \noindent
\textbf{Competitors.} 
We compare \acr{} with Stochastic Concept Bottleneck Models (SCBM) \cite{vandenhirtz2024stochastic} and Causal Concept Graph Models (CGM) \cite{dominici2024causal}, state-of-the-art CBMs developed with a strong focus on intervenability. We also compare with a neural network (NN) directly predicting the concepts, which is the concept predictor for most CBMs, such as Concept Bottleneck Models (CBNM) \cite{koh2020concept}, Concept-based Memory Reasoner \cite{debot2024interpretable}, and Concept Residual Models \cite{mahinpei2021promises}. In the task accuracy experiment, we compare with SCBM, CGM, and CBNM. 

\subsection{Key findings}

\begin{figure}
    \centering
    \includegraphics[width=.99\linewidth]{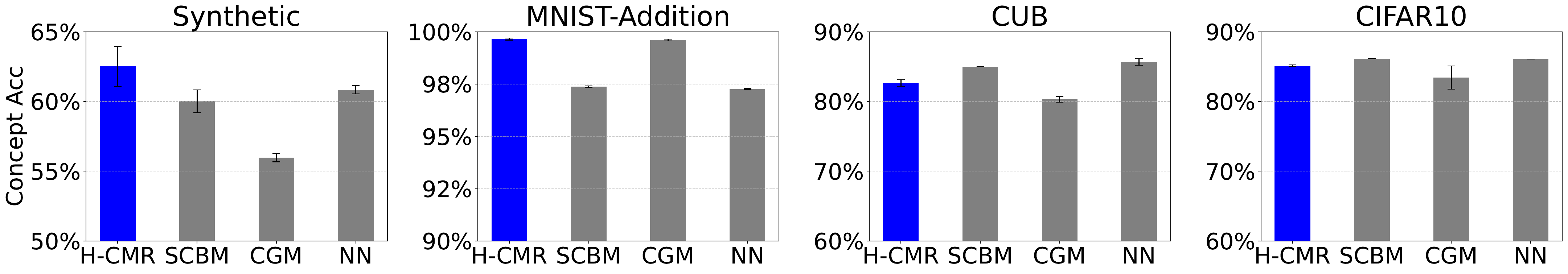}    
    \caption{Concept accuracy for all datasets and models.}
    \label{fig:accuracy}
\end{figure}

\begin{figure}
    \centering
    \begin{minipage}[t]{0.48\linewidth}
        \centering
        \includegraphics[width=.72\linewidth]{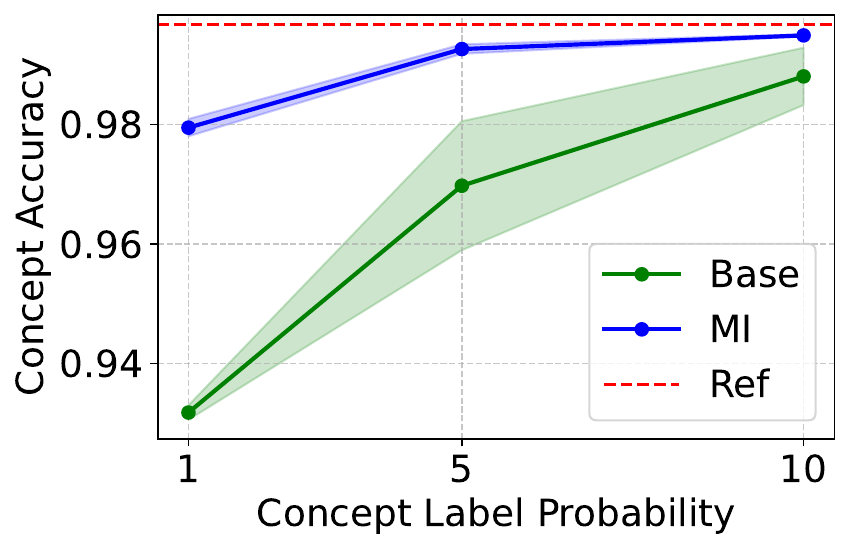}
        \caption{Data efficiency of \acr{} with (MI) and without (Base) background knowledge on MNIST-Add. The x-axis denotes how many concept labels are included in the training set. The reference is accuracy when training on all labels.}
        \label{fig:data_efficiency}
    \end{minipage}
    \hfill
    \begin{minipage}[t]{0.48\linewidth}
        \centering
        \includegraphics[width=.72\linewidth]{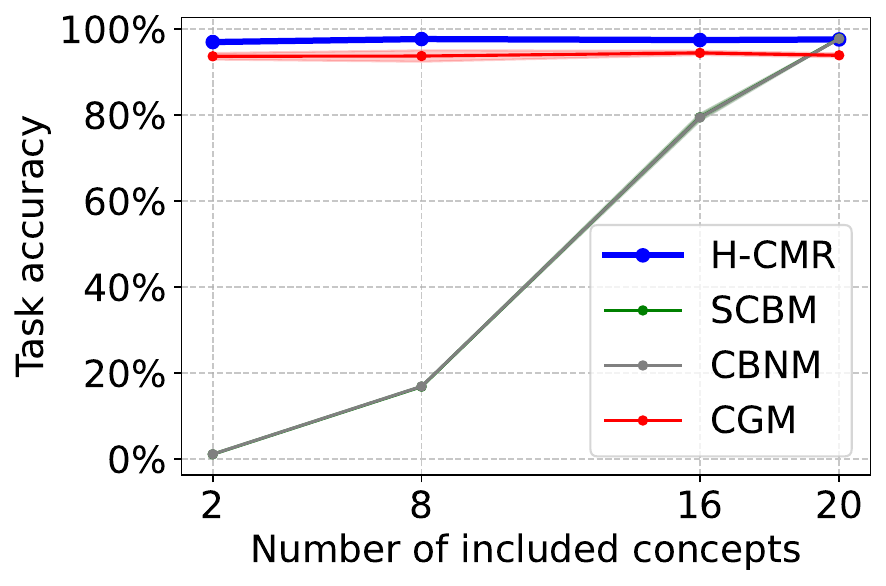}
        \caption{Task accuracy on MNIST-Addition for training on different sizes of the concept set. Universal classifiers (\acr{}, CGM) are robust to the choice of concepts, while other approaches are not (SCBM, CBNM).}
        \label{fig:task_acc}
    \end{minipage}
\end{figure}

\textbf{\acr{}'s interpretability does not harm concept accuracy (Figure \ref{fig:accuracy}), and achieves high task accuracy irrespective of the concept set (Figure \ref{fig:task_acc}).} \acr{} achieves similar levels of concept accuracy compared to competitors. As a universal classifier, \acr{} can achieve high task accuracy even with small concept sets, similar to some other CBMs \cite{EspinosaZarlenga2022cem, mahinpei2021promises, debot2024interpretable}.

\begin{figure}
    \centering
    \includegraphics[width=.9\linewidth]{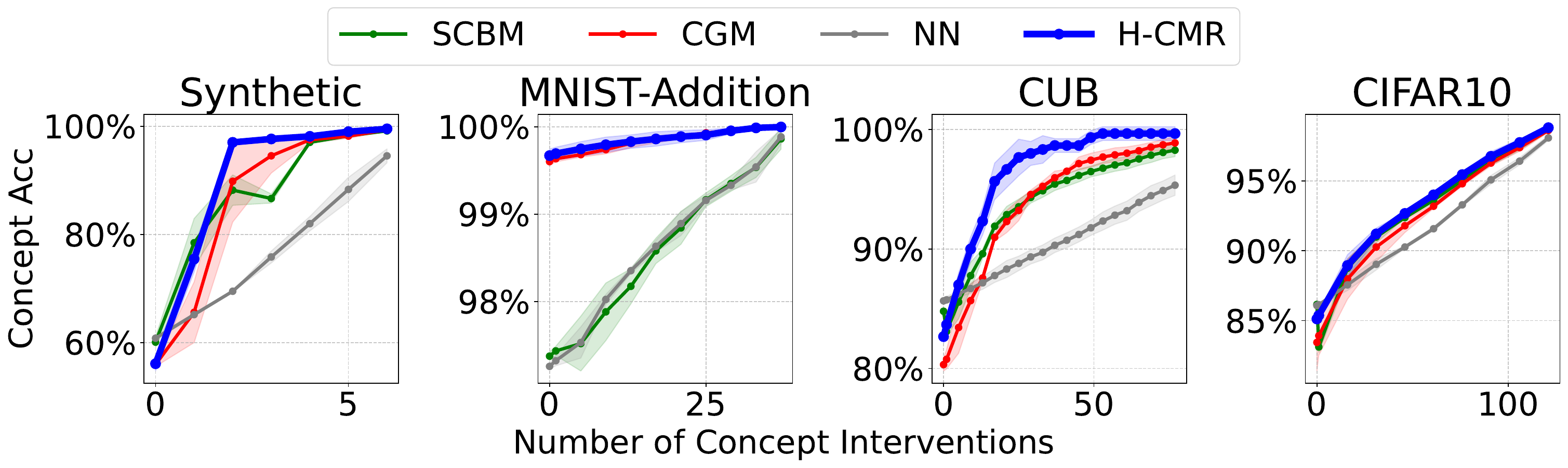}    
    \caption{Concept accuracy before vs.\ after intervening on increasingly more concepts.
    }
    \label{fig:interventions}
\end{figure}

\bigskip \noindent
\textbf{\acr{} shows a high degree of intervenability (Figure \ref{fig:interventions}).} We evaluate \acr{}'s gain in concept accuracy after intervening on increasingly more concepts. 
\acr{} demonstrates far higher degrees of intervenability compared to CBMs modelling concepts independently (NN), and similar or better to approaches that model concepts dependently (SCBM, CGM).

\bigskip \noindent
\textbf{Model interventions by human experts improve data efficiency during training (Figure \ref{fig:data_efficiency}).} We exploit these interventions to provide \acr{} with background knowledge about a subset of the concepts for MNIST-Addition, allowing \acr{} to maintain high accuracy even in low-data regimes with only partial supervision on the concepts. We give \acr{} rules defining 18 concepts in terms of the 10 remaining ones, and force the latter to be source concepts.
This helps in two ways. First, when source concepts are correctly predicted, so are the others. Second, when for a training example a label is only available on non-sources, the gradient can backpropagate through the given rules to provide a training signal to the sources. This means \acr{} does not only work in a concept-based setting, where full concept supervision is typically provided, but also in a neurosymbolic setting, where often \textit{distant supervision} is used to train concepts by exploiting background knowledge \cite{manhaeve2018deepproblog}.

\bigskip \noindent
\textbf{\acr{} learns meaningful rules.} We qualitatively inspect the rules \acr{} learns in Appendix \ref{app:more_results}. For instance, the rules for MNIST-Addition show that \acr{} has learned that there is only one digit in each MNIST image (i.e.\ mutually exclusive concepts).

\section{Related work}

\acr{} is related to two major directions in concept-based models (CBMs) research: one focusing on closing the \textit{accuracy gap} between CBMs and black-box models like deep neural networks \cite{EspinosaZarlenga2022cem, mahinpei2021promises, debot2024interpretable, barbiero2023interpretable}, and one focusing on \textit{intervenability} \citep{dominici2024causal, vandenhirtz2024stochastic, espinosa2023learning, havasi2022addressing}. The former has led to the development of many CBMs that are \textit{universal classifiers}, meaning they can achieve task accuracies comparable to black boxes irrespective of the concept set. However, many of these models achieve this by sacrificing the interpretability of their task predictions \cite{EspinosaZarlenga2022cem, mahinpei2021promises}. A notable exception is Concept-based Memory Reasoner (CMR), a universal classifier which maintains interpretability \cite{debot2024interpretable}. CMR achieves this by modelling the task  as the symbolic execution of a neurally selected logic rule from a learned memory.

\begin{wraptable}{r}{0.35\textwidth}
    \scriptsize
    \centering
    \renewcommand{\arraystretch}{1.3}
    \caption{CBMs having properties (\cmark), partially ($\hmark$) or not at all (\xmark): Universal Classifier (UC), Interpretable Predictions (IP), Expressive concept Interventions (EI), Model Interventions (MI).}
    \begin{tabular}{lcccc}
        \toprule
        \textbf{Model} & UC & 
        IP & 
        EI & 
        MI \\
        \midrule
        CBNM \cite{koh2020concept}   
        & \xmark & $\hmark$ & \xmark        & $\hmark$ \\
        CEM \cite{EspinosaZarlenga2022cem}   
        & \cmark & \xmark & \xmark        & \xmark \\
        CMR \cite{debot2024interpretable}   
        & \cmark & $\hmark$ & \xmark        & $\hmark$ \\
        SCBM \cite{vandenhirtz2024stochastic}
        & \xmark & $\hmark$ & $\hmark$        & \xmark \\
        CGM \cite{dominici2024causal}
        & \cmark & \xmark & \cmark        & $\hmark$ \\ \midrule
        H-CMR
        & \cmark & \cmark & \cmark        & \cmark \\
        \bottomrule
    \end{tabular}
    \vspace{-10pt}
    \label{tab:model_comparison_sub}
\end{wraptable}
\bigskip \noindent
However, the aforementioned CBMs model the concepts as conditionally independent. 
This limits the effects of \textit{concept interventions} at test time, where a human expert corrects some wrongly predicted concepts. In such models, concept interventions cannot affect other (possibly correlated) concepts, only the downstream task in a direct way. To address this, a second line of work has emerged \cite{dominici2024causal, vandenhirtz2024stochastic, havasi2022addressing}. 
For instance, Stochastic Concept Bottleneck Models (SCBMs) jointly model concepts rather than treating them as separate variables \cite{vandenhirtz2024stochastic}, allowing interventions to affect other concepts as well. However, SCBMs are not universal classifiers, meaning that their task accuracy depends on the employed concept set.
Causal Concept Graph Models (CGMs) improve intervenability and are universal classifiers \cite{dominici2024causal}. CGMs learn a graph over concepts and apply black-box message-passing, which allows for highly intervenable and expressive predictions. However, CGMs lack interpretability for concepts and tasks due to the black-box message passing.

\bigskip \noindent
Our model, \acr{}, can be seen as an extension of CMR’s symbolic reasoning approach to concept predictions, combining it with CGMs’ idea of concept graph learning. \acr{} achieves expressive interventions, and unlike previous models, it is a universal classifier that provides interpretability at both the concept and task levels. Table \ref{tab:model_comparison_sub} gives an overview of these CBMs.

\bigskip \noindent
We are also related to neurosymbolic approaches that perform rule learning. Some approaches operate on structured relational data such as knowledge graphs, where target predicates are predefined and a perception component is typically absent \cite{cheng2022rlogic, qu2020rnnlogic}. Others resemble standard CBMs, where the model’s structure, i.e.\ which symbols ("tasks") are predicted from which others ("concepts"), is manually defined by the user \cite{si2019synthesizing, daniele2022deep, tang2023perception}. In contrast, \acr{} learns both the symbolic rules and the dependency structure, a directed graph that defines how concepts and tasks depend on each other. Moreover, such works typically do not provide formal guarantees on expressivity, whereas we prove that H-CMR is a universal classifier.

\section{Conclusion}

We introduce \acr{}, a concept-based model that is a universal binary classifier while providing interpretability for both concept and task prediction. Through our experiments, we show that \acr{} (1) achieves state-of-the-art concept and task accuracy, (2) is highly responsive to concept interventions at inference time, and (3) that through model interventions, background knowledge can be incorporated to improve data efficiency, if available. 
H-CMR can have societal impact by improving transparency and human-AI interaction.

\bigskip \noindent
\textbf{Limitations and future work.} Interesting directions for future work include extending the intervention strategy to take uncertainty into account, and performing a more extensive investigation of \acr{}'s performance in a hybrid setting between concept-based and neurosymbolic, where for some concepts expert knowledge is available, and for others concept supervision.

\section*{Acknowledgements}

\sloppy 
This research has received funding from the KU Leuven Research Fund (GA No.\ STG/22/021, CELSA/24/008), from the Research Foundation-Flanders FWO (GA No.\ 1185125N, G033625N),  from the Flemish Government under the “Onderzoeksprogramma Artificiële Intelligentie (AI) Vlaanderen” programme and from the European Union’s Horizon Europe research and innovation programme under the MSCA grant agreement No 10107330. PB acknowledges support from the Swiss National Science Foundation project IMAGINE (No.\ 224226). GD acknowledges support from the European Union’s Horizon Europe project SmartCHANGE (No.\ 101080965), TRUST-ME (No.\ 205121L\_214991) and from the Swiss National Science Foundation projects XAI-PAC (No.\ PZ00P2\_216405).

\bibliographystyle{unsrtnat}
\bibliography{related}

\begin{thebibliography}{33}
\providecommand{\natexlab}[1]{#1}
\providecommand{\url}[1]{\texttt{#1}}
\expandafter\ifx\csname urlstyle\endcsname\relax
  \providecommand{\doi}[1]{doi: #1}\else
  \providecommand{\doi}{doi: \begingroup \urlstyle{rm}\Url}\fi

\bibitem[Koh et~al.(2020)Koh, Nguyen, Tang, Mussmann, Pierson, Kim, and
  Liang]{koh2020concept}
Pang~Wei Koh, Thao Nguyen, Yew~Siang Tang, Stephen Mussmann, Emma Pierson, Been
  Kim, and Percy Liang.
\newblock Concept bottleneck models.
\newblock In \emph{International conference on machine learning}, pages
  5338--5348. PMLR, 2020.

\bibitem[Alvarez~Melis and Jaakkola(2018)]{alvarez2018towards}
David Alvarez~Melis and Tommi Jaakkola.
\newblock Towards robust interpretability with self-explaining neural networks.
\newblock \emph{Advances in neural information processing systems}, 31, 2018.

\bibitem[Chen et~al.(2020)Chen, Bei, and Rudin]{chen2020concept}
Zhi Chen, Yijie Bei, and Cynthia Rudin.
\newblock Concept whitening for interpretable image recognition.
\newblock \emph{Nature Machine Intelligence}, 2\penalty0 (12):\penalty0
  772--782, 2020.

\bibitem[Espinosa~Zarlenga et~al.(2022)Espinosa~Zarlenga, Barbiero, Ciravegna,
  Marra, Giannini, Diligenti, Shams, Precioso, Melacci, Weller, Lio, and
  Jamnik]{EspinosaZarlenga2022cem}
Mateo Espinosa~Zarlenga, Pietro Barbiero, Gabriele Ciravegna, Giuseppe Marra,
  Francesco Giannini, Michelangelo Diligenti, Zohreh Shams, Frederic Precioso,
  Stefano Melacci, Adrian Weller, Pietro Lio, and Mateja Jamnik.
\newblock Concept embedding models: Beyond the accuracy-explainability
  trade-off.
\newblock \emph{Advances in Neural Information Processing Systems}, 35, 2022.

\bibitem[Mahinpei et~al.(2021)Mahinpei, Clark, Lage, Doshi-Velez, and
  Pan]{mahinpei2021promises}
Anita Mahinpei, Justin Clark, Isaac Lage, Finale Doshi-Velez, and Weiwei Pan.
\newblock Promises and pitfalls of black-box concept learning models.
\newblock \emph{arXiv preprint arXiv:2106.13314}, 2021.

\bibitem[Debot et~al.(2024)Debot, Barbiero, Giannini, Ciravegna, Diligenti, and
  Marra]{debot2024interpretable}
David Debot, Pietro Barbiero, Francesco Giannini, Gabriele Ciravegna,
  Michelangelo Diligenti, and Giuseppe Marra.
\newblock Interpretable concept-based memory reasoning.
\newblock \emph{Advances of neural information processing systems 37, NeurIPS
  2024}, 2024.

\bibitem[Barbiero et~al.(2023)Barbiero, Ciravegna, Giannini, Zarlenga,
  Magister, Tonda, Lio', Precioso, Jamnik, and
  Marra]{barbiero2023interpretable}
Pietro Barbiero, Gabriele Ciravegna, Francesco Giannini, Mateo~Espinosa
  Zarlenga, Lucie~Charlotte Magister, Alberto Tonda, Pietro Lio', Frederic
  Precioso, Mateja Jamnik, and Giuseppe Marra.
\newblock Interpretable neural-symbolic concept reasoning.
\newblock In \emph{ICML}, 2023.

\bibitem[Poeta et~al.(2023)Poeta, Ciravegna, Pastor, Cerquitelli, and
  Baralis]{poeta2023concept}
Eleonora Poeta, Gabriele Ciravegna, Eliana Pastor, Tania Cerquitelli, and Elena
  Baralis.
\newblock Concept-based explainable artificial intelligence: A survey.
\newblock \emph{arXiv preprint arXiv:2312.12936}, 2023.

\bibitem[Dominici et~al.(2024)Dominici, Barbiero, Zarlenga, Termine, Gjoreski,
  Marra, and Langheinrich]{dominici2024causal}
Gabriele Dominici, Pietro Barbiero, Mateo~Espinosa Zarlenga, Alberto Termine,
  Martin Gjoreski, Giuseppe Marra, and Marc Langheinrich.
\newblock Causal concept graph models: Beyond causal opacity in deep learning.
\newblock \emph{arXiv preprint arXiv:2405.16507}, 2024.

\bibitem[Vandenhirtz et~al.(2024)Vandenhirtz, Laguna, Marcinkevi{\v{c}}s, and
  Vogt]{vandenhirtz2024stochastic}
Moritz Vandenhirtz, Sonia Laguna, Ri{\v{c}}ards Marcinkevi{\v{c}}s, and Julia
  Vogt.
\newblock Stochastic concept bottleneck models.
\newblock \emph{Advances in Neural Information Processing Systems},
  37:\penalty0 51787--51810, 2024.

\bibitem[Espinosa~Zarlenga et~al.(2023)Espinosa~Zarlenga, Collins, Dvijotham,
  Weller, Shams, and Jamnik]{espinosa2023learning}
Mateo Espinosa~Zarlenga, Katie Collins, Krishnamurthy Dvijotham, Adrian Weller,
  Zohreh Shams, and Mateja Jamnik.
\newblock Learning to receive help: Intervention-aware concept embedding
  models.
\newblock \emph{Advances in Neural Information Processing Systems},
  36:\penalty0 37849--37875, 2023.

\bibitem[Havasi et~al.(2022)Havasi, Parbhoo, and
  Doshi-Velez]{havasi2022addressing}
Marton Havasi, Sonali Parbhoo, and Finale Doshi-Velez.
\newblock Addressing leakage in concept bottleneck models.
\newblock \emph{Advances in Neural Information Processing Systems},
  35:\penalty0 23386--23397, 2022.

\bibitem[Hornik et~al.(1989)Hornik, Stinchcombe, and
  White]{hornik1989multilayer}
Kurt Hornik, Maxwell Stinchcombe, and Halbert White.
\newblock Multilayer feedforward networks are universal approximators.
\newblock \emph{Neural networks}, 2\penalty0 (5):\penalty0 359--366, 1989.

\bibitem[Massidda et~al.(2023)Massidda, Landolfi, Cinquini, and
  Bacciu]{massidda2023constraint}
Riccardo Massidda, Francesco Landolfi, Martina Cinquini, and Davide Bacciu.
\newblock Constraint-free structure learning with smooth acyclic orientations.
\newblock \emph{arXiv preprint arXiv:2309.08406}, 2023.

\bibitem[Marconato et~al.(2022)Marconato, Passerini, and
  Teso]{marconato2022glancenets}
Emanuele Marconato, Andrea Passerini, and Stefano Teso.
\newblock Glancenets: Interpretable, leak-proof concept-based models.
\newblock \emph{Advances in Neural Information Processing Systems},
  35:\penalty0 21212--21227, 2022.

\bibitem[Welinder et~al.(2010)Welinder, Branson, Mita, Wah, Schroff, Belongie,
  and Perona]{cub}
Peter Welinder, Steve Branson, Takeshi Mita, Catherine Wah, Florian Schroff,
  Serge Belongie, and Pietro Perona.
\newblock Caltech-ucsd birds 200.
\newblock Technical Report CNS-TR-201, Caltech, 2010.
\newblock URL \url{/se3/wp-content/uploads/2014/09/WelinderEtal10_CUB-200.pdf,
  http://www.vision.caltech.edu/visipedia/CUB-200.html}.

\bibitem[Manhaeve et~al.(2018)Manhaeve, Dumancic, Kimmig, Demeester, and
  Raedt]{manhaeve2018deepproblog}
Robin Manhaeve, Sebastijan Dumancic, Angelika Kimmig, Thomas Demeester, and
  Luc~De Raedt.
\newblock {DeepProbLog: Neural Probabilistic Logic Programming}.
\newblock In \emph{NeurIPS}, pages 3753--3763, 2018.

\bibitem[Lecun et~al.(1998)Lecun, Bottou, Bengio, and Haffner]{mnistLecun}
Y.~Lecun, L.~Bottou, Y.~Bengio, and P.~Haffner.
\newblock Gradient-based learning applied to document recognition.
\newblock \emph{Proceedings of the IEEE}, 86\penalty0 (11):\penalty0
  2278--2324, 1998.
\newblock \doi{10.1109/5.726791}.

\bibitem[Krizhevsky et~al.(2009)Krizhevsky, Hinton,
  et~al.]{krizhevsky2009learning}
Alex Krizhevsky, Geoffrey Hinton, et~al.
\newblock Learning multiple layers of features from tiny images.(2009), 2009.

\bibitem[Oikarinen et~al.(2023)Oikarinen, Das, Nguyen, and
  Weng]{oikarinen2023labelfree}
Tuomas Oikarinen, Subhro Das, Lam~M. Nguyen, and Tsui-Wei Weng.
\newblock Label-free concept bottleneck models, 2023.

\bibitem[Cheng et~al.(2022)Cheng, Liu, Wang, and Sun]{cheng2022rlogic}
Kewei Cheng, Jiahao Liu, Wei Wang, and Yizhou Sun.
\newblock Rlogic: Recursive logical rule learning from knowledge graphs.
\newblock In \emph{Proceedings of the 28th ACM SIGKDD conference on knowledge
  discovery and data mining}, pages 179--189, 2022.

\bibitem[Qu et~al.(2020)Qu, Chen, Xhonneux, Bengio, and Tang]{qu2020rnnlogic}
Meng Qu, Junkun Chen, Louis-Pascal Xhonneux, Yoshua Bengio, and Jian Tang.
\newblock Rnnlogic: Learning logic rules for reasoning on knowledge graphs.
\newblock \emph{arXiv preprint arXiv:2010.04029}, 2020.

\bibitem[Si et~al.(2019)Si, Raghothaman, Heo, and Naik]{si2019synthesizing}
Xujie Si, Mukund Raghothaman, Kihong Heo, and Mayur Naik.
\newblock Synthesizing datalog programs using numerical relaxation.
\newblock \emph{arXiv preprint arXiv:1906.00163}, 2019.

\bibitem[Daniele et~al.(2022)Daniele, Campari, Malhotra, and
  Serafini]{daniele2022deep}
Alessandro Daniele, Tommaso Campari, Sagar Malhotra, and Luciano Serafini.
\newblock Deep symbolic learning: Discovering symbols and rules from
  perceptions.
\newblock \emph{arXiv preprint arXiv:2208.11561}, 2022.

\bibitem[Tang and Ellis(2023)]{tang2023perception}
Hao Tang and Kevin Ellis.
\newblock From perception to programs: regularize, overparameterize, and
  amortize.
\newblock In \emph{International Conference on Machine Learning}, pages
  33616--33631. PMLR, 2023.

\bibitem[LeCun et~al.(1998)LeCun, Bottou, Bengio, and Haffner]{lecun1998mnist}
Yann LeCun, L{\'e}on Bottou, Yoshua Bengio, and Patrick Haffner.
\newblock Gradient-based learning applied to document recognition.
\newblock \emph{IEEE}, 86\penalty0 (11):\penalty0 2278--2324, 1998.

\bibitem[He et~al.(2016)He, Zhang, Ren, and Sun]{he2016deep}
Kaiming He, Xiangyu Zhang, Shaoqing Ren, and Jian Sun.
\newblock Deep residual learning for image recognition.
\newblock In \emph{Proceedings of the IEEE conference on computer vision and
  pattern recognition}, pages 770--778, 2016.

\bibitem[Rosch(1978)]{rosch1978principles}
Eleanor Rosch.
\newblock Principles of categorization.
\newblock In \emph{Cognition and categorization}, pages 27--48. Routledge,
  1978.

\bibitem[Rudin(2019)]{rudin2019stop}
Cynthia Rudin.
\newblock Stop explaining black box machine learning models for high stakes
  decisions and use interpretable models instead.
\newblock \emph{Nature machine intelligence}, 1\penalty0 (5):\penalty0
  206--215, 2019.

\bibitem[Li et~al.(2018)Li, Liu, Chen, and Rudin]{li2018deep}
Oscar Li, Hao Liu, Chaofan Chen, and Cynthia Rudin.
\newblock Deep learning for case-based reasoning through prototypes: A neural
  network that explains its predictions.
\newblock In \emph{Proceedings of the AAAI Conference on Artificial
  Intelligence}, volume~32, 2018.

\bibitem[Chen et~al.(2019)Chen, Li, Tao, Barnett, Rudin, and Su]{chen2019looks}
Chaofan Chen, Oscar Li, Daniel Tao, Alina Barnett, Cynthia Rudin, and
  Jonathan~K Su.
\newblock This looks like that: deep learning for interpretable image
  recognition.
\newblock \emph{Advances in neural information processing systems}, 32, 2019.

\bibitem[Paszke et~al.(2019)Paszke, Gross, Massa, Lerer, Bradbury, Chanan,
  Killeen, Lin, Gimelshein, Antiga, Desmaison, Köpf, Yang, DeVito, Raison,
  Tejani, Chilamkurthy, Steiner, Fang, Bai, and Chintala]{paszke2019pytorch}
Adam Paszke, Sam Gross, Francisco Massa, Adam Lerer, James Bradbury, Gregory
  Chanan, Trevor Killeen, Zeming Lin, Natalia Gimelshein, Luca Antiga, Alban
  Desmaison, Andreas Köpf, Edward Yang, Zach DeVito, Martin Raison, Alykhan
  Tejani, Sasank Chilamkurthy, Benoit Steiner, Lu~Fang, Junjie Bai, and Soumith
  Chintala.
\newblock Pytorch: An imperative style, high-performance deep learning library,
  2019.

\bibitem[Pedregosa et~al.(2011)Pedregosa, Varoquaux, Gramfort, Michel, Thirion,
  Grisel, Blondel, Prettenhofer, Weiss, Dubourg, Vanderplas, Passos,
  Cournapeau, Brucher, Perrot, and Duchesnay]{scikit-learn}
F.~Pedregosa, G.~Varoquaux, A.~Gramfort, V.~Michel, B.~Thirion, O.~Grisel,
  M.~Blondel, P.~Prettenhofer, R.~Weiss, V.~Dubourg, J.~Vanderplas, A.~Passos,
  D.~Cournapeau, M.~Brucher, M.~Perrot, and E.~Duchesnay.
\newblock Scikit-learn: Machine learning in {P}ython.
\newblock \emph{Journal of Machine Learning Research}, 12:\penalty0 2825--2830,
  2011.

\end{thebibliography}

\newpage

\newpage

\appendix
{\huge \textbf{Supplementary Material}}

\addtocontents{toc}{\protect\setcounter{tocdepth}{2}}
\renewcommand{\contentsname}{\large Table of Contents}
\tableofcontents
\newpage

\section{Details of properties in \acr{}} \label{app:props}

\subsection{Task prediction}

In \acr{}, tasks and concepts are modelled in the same way: they are nodes that are predicted from their parent nodes, and the learned rules in the memory define this structure. This means that the memory contains rules for predicting each concept, and for predicting each task. Consequently, the parametrization explained in the main text, which defines how concepts are predicted from other concepts, is also used for tasks. The simplest way to incorporate tasks is by simply considering them as additional concepts. Then, \acr{} learns a graph over concepts and tasks. This allows for instance that tasks are predicted using each other, and that concepts are predicted using tasks. 

\bigskip \noindent
This approach is also possible in e.g.\ CGM \cite{dominici2024causal}, but not in most CBMs, where concepts and tasks are modelled in two separate layers of the model. In most CBMs, concepts are first predicted from the input using a neural network, and then the task is predicted from the concepts (e.g.\ CBNM \cite{koh2020concept}) and possibly some residual, e.g.\ an embedding to provide additional contextual information \cite{mahinpei2021promises}. 

\bigskip \noindent
In \acr{}, this concept-task structure can be enforced by forcing the tasks to be sink nodes in the learned graph, i.e.\ they have no outgoing edges (no concepts are predicted from the tasks, and tasks are not predicted using each other), that additionally have all concepts as potential parents. This can be done through model interventions.

\subsection{Model interventions}

In this section, we give some examples on how human experts can do model interventions on \acr{} during training, influencing the model that is being learned. To this end, we first derive the following matrix $A \in \mathbb{R}^{n_C \times n_C}$ from \acr{}'s parametrization:
\begin{align}
    \forall i,j \in [1,n_C]: A_{ij} = \mathbbm{1}[O_j > O_i]
\end{align}
where $A_{ij}=1$ indicates that concept $j$ is allowed to be a parent of concept $i$ based on the node priority vector $O$ (see Equation \ref{eq:true_roles}). This matrix serves as an alternative representation of the parent-child constraints originally encoded by $O$, which can then be used in Equation \ref{eq:true_roles} instead of $O$. By intervening on this matrix, it is possible to:
\begin{itemize}
    \item force a concept $k$ to be a source (no parents): set $ \forall k \in [1, n_C]: A_{kj} := 0$;
    \item force a concept $k$ to be a sink (no children): set $ \forall k \in [1, n_C]: A_{jk} := 0$;
    \item forbid a concept $m$ to be a parent of concept $k$: set $A_{km} := 0$.
\end{itemize}
Furthermore, a specific topological ordering of the concepts can be enforced by explicitly assigning values to the entire node priority vector $O$. Moreover, to ensure that a concept $l$ precedes or follows concept $k$ in the topological ordering, one can set $O_l := O_k - z$ or $O_l := O_k + z$, respectively, where $z \in \mathbb{R}^+_0$ is any chosen positive number.

\bigskip \noindent
The human expert can also intervene on the roles the concepts play in individual rules. This means intervening on the 'unadjusted roles' $R'$, which are combined with the node priorities $O$ to form the rules. For instance, by intervening on $R'$ (and possibly $O$), it is possible to:
\begin{itemize}
    \item force a concept $k$ to be absent in rule $l$ of concept $i$: set $p(R'_{i,l,k} = I):=1$, $p(R'_{i,l,k} = P):=0$, and $p(R'_{i,l,k} = N):=0$;
    \item force a concept $k$ to be positively present in rule $l$ of concept $i$ (assuming $O$ allows it): set $p(R'_{i,l,k} = I):=0$, $p(R'_{i,l,k} = P):=1$, and $p(R'_{i,l,k} = N):=0$;
    \item force a concept $k$ to be negatively present in rule $l$ of concept $i$ (assuming $O$ allows it): set $p(R'_{i,l,k} = I):=0$, $p(R'_{i,l,k} = P):=0$, and $p(R'_{i,l,k} = N):=1$.
\end{itemize}
By intervening on the roles and the node priorities in these ways, experts can have fine-grained control over the content and structure of the rules. In the extreme case, human experts can choose to fully specify a rule, or even the entire rule set, through such interventions.

\subsection{Verification}

Since the (learned) memory of rules is transparent, it can be formally verified against desired constraints in a similar fashion as for CMR \cite{debot2024interpretable}. For instance, one can verify whether a constraint such as "\textit{whenever the concept 'black wings' is predicted as True, the concept 'white wings' is predicted as False and the task 'pigeon' is predicted as False}" is guaranteed by the learned rules. This is possible because both concepts and tasks predictions can be represented as disjunctions over the rules in the memory, expressed in propositional logic. As described by \citet{debot2024interpretable}, the neural rule selection can be encoded within this disjunction by introducing additional propositional atoms that denote whether each individual rule is selected, along with mutual exclusivity constraints between these atoms.
Consequently, standard formal verification tools (e.g.\ model checkers) can be employed to verify constraints w.r.t.\ this propositional logic formula. We refer to Section 4.3 of \citet{debot2024interpretable}.

\section{Experimental and implementation details}  \label{app:exp_details}

\textbf{Datasets.} In CUB \cite{cub}, there are 112 concepts related to bird characteristics, such as wing pattern and head size. Each input consists of a single image containing a bird. In MNIST-Addition \cite{manhaeve2018deepproblog}, the input consists of two MNIST images \cite{lecun1998mnist}. There are 10 concepts per image, representing the digit present, and 19 tasks corresponding to the possible sums of the two digits. For CIFAR10, which does not include predefined concepts, we use the same technique as \citet{oikarinen2023labelfree} to obtain them. Specifically, we use the same concept set as \citet{oikarinen2023labelfree}, which they obtained by prompting an LLM, and obtain concept annotations by exploiting vision-language models, as in their work. For our synthetic dataset, we modify MNIST-Addition by restricting it to examples containing only the digits zero and one. We discard the original concepts and tasks and instead generate new concepts and their corresponding labels for each example using the following sampling process:
\begin{itemize}
    \item $p(C_0=1) = \begin{cases}
        0.7 & \text{if the first digit is a 1} \\
        0  & \text{otherwise}
    \end{cases}$
    \item $p(C_1=1) = \begin{cases}
        0.7 & \text{if the second digit is a 1} \\
        0  & \text{otherwise}
    \end{cases}$
    \item $p(C_2=1 \mid \hat{c}_0, \hat{c}_1) = \hat{c}_0 \oplus' \hat{c}_1$
    \item $p(C_3=1 \mid \hat{c}_0, \hat{c}_2) = \hat{c}_0 \oplus' \hat{c}_2$
    \item $p(C_4=1 \mid \hat{c}_1, \hat{c}_2) = \hat{c}_1 \oplus' \hat{c}_2$
    \item $p(C_5=1 \mid \hat{c}_3, \hat{c}_4) = \hat{c}_3 \oplus' \hat{c}_4$
    \item $p(C_6=1 \mid \hat{c}_0, \hat{c}_1) = \hat{c}_0 \oplus' \hat{c}_1$
\end{itemize}
where $\oplus$ is the logical XOR, and where we define $\oplus'$ as a noisy XOR:
\begin{align}
    \hat{c}_i \oplus' \hat{c}_j = \begin{cases}
        1 & \text{if} \,\, \hat{c}_i \oplus \hat{c}_j = 1 \\
        0.05  & \text{otherwise}
    \end{cases}
\end{align}
For each example, we sample labels from the above distributions. Intuitively, the concepts $C_0$ and $C_1$ indicate whether the corresponding MNIST images contain the digit one; however, these labels are intentionally noisy. The remaining concepts are constructed as noisy logical XORs of $C_0$, $C_1$, and of each other, introducing additional complexity and interdependence among the concepts.

\bigskip \noindent
\textbf{Reproducibility.} For reproducibility, we used seeds 0, 1 and 2 in all experiments.

\bigskip \noindent
\textbf{Model input.} For MNIST-Addition, CIFAR10 and the synthetic dataset, we train directly on the images. For CIFAR10, we use the same setup as \citet{vandenhirtz2024stochastic}.
For CUB, we instead use pretrained Resnet18 embeddings \cite{he2016deep}, using the setup of \citet{debot2024interpretable}.

\bigskip \noindent
\textbf{General training information.} We use the AdamW optimizer. For \acr{} and CBNMs, we maximize the likelihood of the data. SCBMs and CGMs are trained using their custom loss functions. After training, we select the model checkpoint with the highest validation accuracy. Validation accuracy refers to concept prediction accuracy in all experiments, except in the MNIST-Addition setting where tasks are retained. In that case, accuracy is computed over the concatenation of both concepts and tasks. Throughout, we model all concepts and tasks as independent Bernoulli random variables.

\bigskip \noindent
\textbf{Intervention policy.} For the results presented in the main text, the intervention policy follows the graph learned by \acr{}, intervening first on the sink nodes and gradually moving down the topological ordering as determined by the learned node priority vector.
This approach makes it easy to interpret the results, as the intervention order is the same for different models (i.e.\ if we intervene on 3 concepts, they are the same concepts for \acr{} and all competitors). This makes it clear how well the models are able to \textit{propagate} the intervention to other concept predictions.
To ensure a fair comparison with CGM, we make sure that CGM learns using the same graph that \acr{} learned. Additional ablation studies using different intervention policies are provided in Appendix \ref{app:more_results}.

\bigskip \noindent
\textbf{General architectural details.} 
For each experiment, we define a neural network $\phi$ that maps the input to some latent embedding with size $size_{latent}$ a hyperparameter. The architecture of this neural network depends on the experiment but is the same for all models.

\bigskip \noindent
In \acr{}'s encoder, $f_i$ (see Equation \ref{eq:root_concept_node}) is implemented as first applying $\phi$ to the input, producing a latent embedding with size $size_{latent}$. This passes through a linear layer with leaky ReLU activation outputting 2 embeddings of size $size_{c \, emb}$ per concept (similar to concept embeddings \cite{EspinosaZarlenga2022cem}, we call one embedding the "positive" one, and the other one the "negative" one), with $size_{c \, emb}$ a hyperparameter. For each concept, the two embeddings are concatenated and the result passes through a linear layer with sigmoid activation to produce the concept prediction probability (corresponding to $f_i$ in Equation \ref{eq:root_concept_node}. The concatenation of all embeddings form the output embedding of the encoder ($\hat{e}$ as produced by $g$ in Equation \ref{eq:root_concept_node}).

\bigskip \noindent
For \acr{}'s decoder, the neural selection $p(S_i=k \mid \hat{e}, \hat{c}_{parents(i)})$ is implemented in the following way. First, we mask within $\hat{e}$ the values that originally corresponded to non-parent concepts. Then, for each parent concept, if it is predicted to be True, we mask the values of $\hat{e}$ that correspond to its negative embedding. If it is predicted to be False, we mask the ones corresponding to its positive embedding. Note again that this is similar to concept embeddings \cite{EspinosaZarlenga2022cem}. Next, we concatenate the embedding with the concept predictions $\hat{c}$. For non-parents, we set their value always to 0. The result is a tensor of shape $2\cdot n_c \cdot size_{c \, emb} + n_C$, which is passed through a linear layer (ReLU activation) with output size $size_{latent}$. This is then passed through another linear layer with output size $n_C \cdot n_R$, which is reshaped to the shape $(n_C,n_R)$. For each row $i$ in this tensor, this represents the logits of $p(S_i \mid \cdot)$. We apply a softmax over the last dimension to get the corresponding probabilities.

\bigskip \noindent
In \acr{}'s memory, the node priority vector $O$ is implemented as a torch Embedding of shape $(1, n_C)$. The rule embeddings in the memory (each representing the latent representation of a logic rule) are implemented as a torch Embedding of shape $(n_R \cdot n_C, size_{rule \, emb})$ with $size_{rule \, emb}$ a hyperparameter. This is reshaped to shape $(n_C, n_R, size_{rule \, emb})$. The "rule decoding" neural network consists of a linear layer (leaky ReLU) with output size $size_{rule \, emb}$ followed by a linear layer with output size $n_C \cdot n_R$. After passing the embedding through this neural network, the result has output shape $(n_C, n_R, n_C \cdot 3)$, which is reshaped into $(n_C, n_R, n_C, 3)$. This corresponds to the logits of the 'unadjusted roles' $p(R')$. We obtain the probabilities by applying a softmax to the last dimension. 

\bigskip \noindent
The CBNM first applies $\phi$, and then continues with a linear layer with sigmoid activation outputting the concept probabilities. The task predictor is a feed-forward neural network consisting of 2 hidden layers with ReLU activation and dimension 100, and a final linear layer with sigmoid activation. We employ hard concepts to avoid the problem of concept leakage which may harm interpretability \cite{marconato2022glancenets}, meaning we threshold the concept predictions at 50\% before passing them to the task predictor.

\bigskip \noindent
For SCBM, we use the implementation as given by the authors.\footnote{\url{https://github.com/mvandenhi/SCBM}} The only differences are that (1) we use $\phi$ to first produce a latent embedding which we pass to SCBM, and (2) we replace the softmax on their task prediction by a sigmoid, as we treat the tasks in all experiments independently. For the $\alpha$ hyperparameter, we always use $0.99$. We always use the amortized variant of SCBMs, which is encouraged by \citet{vandenhirtz2024stochastic}. We use $100$ Monte Carlo samples.

\bigskip \noindent
For CGM, we also use the implementation as given by the authors.\footnote{\url{https://github.com/gabriele-dominici/CausalCGM}} The only difference is that we use $\phi$ to first produce a latent embedding, which is passed to CGM. To ensure a fair comparison when measuring intervenability, we equip CGM with the same graph that \acr{} learned.

\bigskip \noindent
\textbf{Hyperparameters per experiment.} 
In CUB, $\phi$ is a feed-forward neural network with 3 hidden layers. Each layer has output size $size_{latent}$. We use a learning rate of 0.001, batch size 1048, and train for 500 epochs. We check validation accuracy every 20 epochs.
For \acr{}, we use $size_{latent}=256$, $size_{rule \, emb}=500$, $size_{c \, emb}=10$, $n_R=5$, and $\beta=0.1$.
For SCBM, we use $size_{latent}=64$.
For CGM, we use $size_{latent}=64$ and $size_{c \, emb}=8$.

\bigskip \noindent
In the synthetic dataset, $\phi$ consists of a convolutional neural network (CNN) consisting of a Conv2d layer (1 input channel, 6 output channels and kernel size 5), a MaxPool2d layer (kernel size and stride 2), a ReLU activation, a Conv2D layer (16 output channels and kernel size 5), another MaxPool2d layer (kernel size and stride 2), another ReLU activation, a flattening layer and finally a linear layer with output size $size_{latent} / 2$. $\phi$ applies this CNN to both input images and concatenates the resulting embeddings. We use a learning rate of 0.001, batch size 256, and train for 100 epochs. We check validation accuracy every 5 epochs.
For \acr{}, we use $size_{latent}=128$, $size_{rule \, emb}=1000$, $size_{c \, emb}=3$, $n_R=10$, and $\beta=0.1$.
For SCBM, we use $size_{latent}=128$.
For CGM, we use $size_{latent}=128$ and $size_{c \, emb}=3$.
For CBNM, we use $size_{latent}=128$.

\bigskip \noindent
In MNIST-Addition, $\phi$ consists of a CNN that is the same as for the synthetic dataset, but with 3 additional linear layers at the end with output size $size_{latent} /2$, the first two having ReLU activation. $\phi$ applies this CNN to both input images and concatenates the resulting embeddings. We use a learning rate of 0.001, batch size 256, and train for 300 epochs. We check validation accuracy every 5 epochs.
For \acr{}, we use $size_{latent}=128$, $size_{rule \, emb}=1000$, $size_{c \, emb}=3$, $n_R=10$, and $\beta=0.1$.
For SCBM, we use $size_{latent}=256$.
For CGM, we use $size_{latent}=100$ and $size_{c \, emb}=8$.
For CBNM, we use $size_{latent}=128$.

\bigskip \noindent
In CIFAR10, $\phi$ consists of a CNN consisting of a Conv2d layer (3 input channels, 32 output channels, kernel size 5 and stride 3), ReLU activation, a Conv2d layer (32 input channels, 64 output channels, kernel size 5 and stride 3), ReLU activation, a MaxPool2d layer (kernel size 2), a Dropout layer (probability of 50\%), a flattening layer and a linear layer with output size $size_{latent}$ with ReLu activation. We use a learning rate of 0.001, batch size 100, and train for 300 epochs. We check validation accuracy every 5 epochs.
For \acr{}, we use $size_{latent}=100$, $size_{rule \, emb}=500$, $size_{c \, emb}=3$, $n_R=10$, and $\beta=0.1$.
For SCBM, we use $size_{latent}=100$.
For CGM, we use $size_{latent}=100$ and $size_{c \, emb}=8$.
For CBNM, we use $size_{latent}=100$.

\bigskip \noindent
\textbf{Additional details of the MNIST-Addition experiments.} In the MNIST-Addition experiment where only concept accuracy is reported, we treat the labels for both the individual digits and their sums as concepts. For the experiment reporting task accuracy, we use the more conventional setting, where the digits are considered as concepts, while the sums of digit pairs are treated as tasks. There, we use model interventions to make the tasks sinks in the graph.

\bigskip \noindent
\textbf{Hyperparameter search.}
The hyperparameters for all models were chosen that result in the highest validation accuracy. For $size_{latent}$, we searched within the grid $[32, 64, 100, 128, 200, 256, 512]$, for $size_{c \, emb}$ within $[2, 3, 8, 16]$, and for $size_{rule \, emb}$ within $[100, 500, 1000]$.

\bigskip \noindent
\textbf{Prototypicality regularization.}
Similar to the approach of CMR \cite{debot2024interpretable}, we employ a regularization term that encourages the learned rules to be more \textit{prototypical} of the seen concepts during training. This aligns with standard theories in cognitive science \cite{rosch1978principles}. While this notion of prototypicality has inspired many so-called prototype-based models \cite{rudin2019stop, li2018deep, chen2019looks}, where prototypes are built in the input space, such as images. Just like CMR, \acr{} differs from such models significantly. For instance, \acr{} gives a logical interpretation to prototypes as being logical rules. Moreover, the prototypes are built in the concept space (as opposed to the input space). We refer to \citet{debot2024interpretable} (specifically, Section 4.2) for more such differences. During training, this regularization is present as an additional factor in the decoder, replacing Equation \ref{eq:non_root_concept_node} with
\begin{align}
    p(C_i = 1 \mid \hat{e}, \hat{c}, \hat{r}_i, \hat{y}) = \sum_{k=1}^{n_{R}} \,\,\,\, \underbrace{p(S_i=k\; \mid \hat{e}, \hat{c}_{parents(i)})}_{\mathclap{\substack{\text{neural selection of rule} \, k \\ \text{using parent concepts + emb.}}}} \,\,\,\, \cdot \,\,\,\, \underbrace{l(\hat{c}_{parents(i)}, \hat{r}_{i,k})}_{\mathclap{\substack{\text{evaluation of concept $i$'s rule} \, k \\ \text{using parent concepts}}}} \,\,\,\, \cdot \,\,\,\, \underbrace{p_{reg}(r_{i,k}=\hat{c})^{\beta \cdot \hat{y}})}_{\mathclap{\substack{\text{prototypicality of} \, \\ \text{concept $i$'s rule $k$}}}}
\end{align}
where $\beta$ is a hyperparameter, and
\begin{align}
    p_{reg}(r_{i,k} = \hat{c}) = \prod_{j=1}^{n_C} (0.5 \cdot \mathbbm{1}[r_{i,k,j}=I] + \mathbbm{1}[\hat{c}_j=1] \cdot \mathbbm{1}[r_{i,k,j}=P] + \mathbbm{1}[\hat{c}_j=0] \cdot \mathbbm{1}[r_{i,k,j}=N])
\end{align}
Inuitively, the rule that is selected should not only provide a correct prediction, but also resemble the seen concepts as much as possible. For instance, if a concept is True, the loss prefers to see a positive role, over irrelevance, over a negative one. Like in \cite{debot2024interpretable}, the regularization only affects positive training examples ($\hat{y}=1$).

\section{Probabilistic graphical model}  \label{app:pgm}

\begin{figure}
    \centering
    \hfill
    \begin{subfigure}{0.4\linewidth}
        \centering
        \begin{tikzpicture}[->, >=stealth, node distance=1.2cm]
            \begin{scope}[scale=0.8, transform shape]
            \node (X) [draw, fill=lightgray, circle] {$X$};
            \node (Sj) [draw, circle, below=1.8cm of X, xshift=4.2cm] {$S_i$};
            \node (C) [draw, fill=lightgray, circle, below=1.2cm of X)] {$C_j$};
            \node (R) [draw, circle, below=1.2cm of C, xshift=1.8cm] {$R_{ikj}$};
            \node (Yj) [draw, circle, below=1.2cm of C, xshift=5.8cm] {$C_i$};
            \node (help) [below=0.6cm of R, xshift=-0.8cm] {};
    
            \node[draw, thick, fit=(C)(R)(help), inner sep=0.3cm, label=below:{$j \in \{1..n_C\} \setminus \{i\}$}] {};
            \node[draw, thick, fit=(R), inner sep=0.1cm, label=below:{$k \in \{1..n_R\}$}] {};
    
            \draw[draw=brown, thick] (X) -- (Sj);
            \draw[draw=red, thick] (X) edge[in=90, out=0] (Yj);
            \draw[draw=blue, thick] (C) -- (Yj);
            \draw[draw=brown, thick] (Sj) -- (Yj);
            \draw[draw=blue, thick] (R) -- (Yj);
            \draw[draw=brown, thick] (C) -- (Sj);
            \draw[draw=brown, thick] (R) -- (Sj);
            \end{scope}
        \end{tikzpicture}
        \caption{During training.}
        \label{app_fig:pgm_training}
    \end{subfigure}
    \hfill
    \begin{subfigure}{0.30\linewidth}
        \centering
        \begin{tikzpicture}[->, >=stealth, node distance=1.2cm]
            \begin{scope}[scale=0.8, transform shape]
            \node (X) [draw, fill=lightgray, circle] {$X$};
            \node (Sj) [draw, circle, below=1.8cm of X, xshift=1.8cm] {$S_i$};
            \node (C) [draw, circle, below=1.2cm of X)] {$C_j$};
            \node (Yj) [draw, circle, below=1.2cm of C, xshift=2.8cm] {$C_i$};
    
            \node[draw, thick, fit=(C), inner sep=0.2cm, label=below:{$j \in parents(i)$}] {};
    
            \draw[draw=black, thick] (X) -- (C);
            \draw[draw=brown, thick] (X) -- (Sj);
            \draw[draw=blue, thick] (C) -- (Yj);
            \draw[draw=brown, thick] (Sj) -- (Yj);
            \draw[draw=brown, thick] (C) -- (Sj);
            \end{scope}
        \end{tikzpicture}
        \caption{During inference for each non-source $C_i$.}
        \label{app_fig:pgm_inf_non_source}
    \end{subfigure}
    \hfill
    \begin{subfigure}{0.20\linewidth}
        \centering
        \begin{tikzpicture}[->, >=stealth, node distance=1.2cm]
            \begin{scope}[scale=0.8, transform shape]
            \node (X) [draw, fill=lightgray, circle] {$X$};
            \node (Yj) [draw, circle, below=1.2cm of X] {$C_i$};
        
            \draw[draw=red, thick] (X) -- (Yj);
            \end{scope}
        \end{tikzpicture}
        \caption{During inference for each source $C_i$.}
        \label{app_fig:pgm_inf_source}
    \end{subfigure}
    \hfill
    \caption{Part of the probabilistic graphical model for computing a single node $C_i$. Red edges denote the prediction by the encoder for source concepts; brown edges denote the rule selection; blue edges denote the rule evaluation. Grey nodes are observed. During inference, the roles are always observed and fixed, so we do not write them. The observed roles determine which concepts are parents of $C_i$, and whether $C_i$ is a source. For each parent $C_j$, the black edge is a "nested" (b) or (c), depending on whether $C_j$ is a source.}
    \label{app_fig:pgm}
\end{figure}
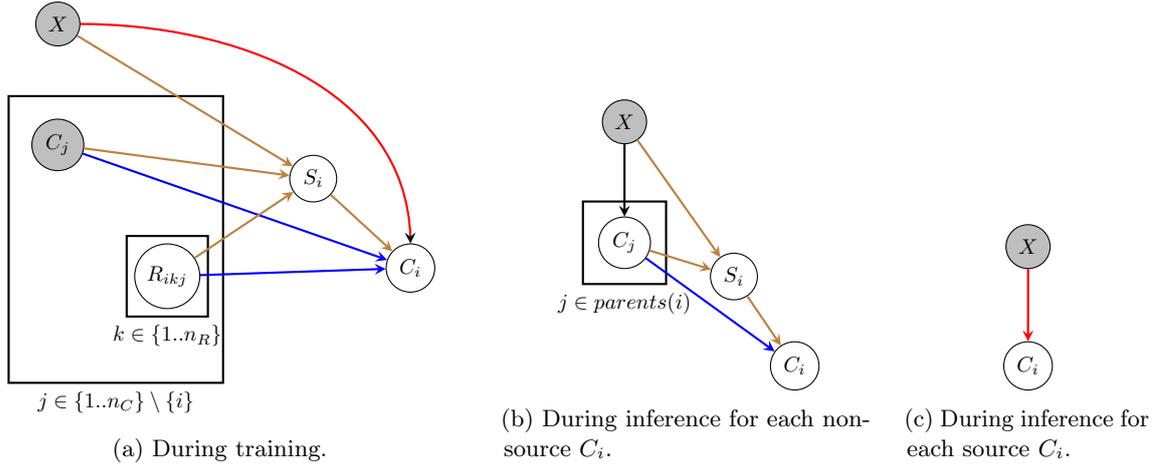

\begin{figure}
    \centering
    \hfill
    \begin{subfigure}{0.9\linewidth}
        \centering
        \begin{tikzpicture}[->, >=stealth, node distance=1.2cm]
            \begin{scope}[scale=0.7, transform shape]
            \node (X) [draw, fill=lightgray, circle] {$X$};
            \node (E) [draw, circle, right=1.2cm of X] {$E$};
            \node (Sj) [draw, circle, below=1.8cm of X, xshift=4.2cm] {$S_i$};
            \node (C) [draw, fill=lightgray, circle, below=1.2cm of X)] {$C_j$};
            \node (Pij) [draw, ellipse, below=3.2cm of X, xshift=1.8cm] {$Parent_{ij}$};
            \node (R) [draw, circle, left=1.2cm of Pij, yshift=-2cm] {$R_{ikj}$};
            \node (Rprime) [draw, circle, left=1.2cm of R] {$R'_{ikj}$};
            \node (O) [draw, circle, below=0.8cm of Rprime] {$O_j$};
            \node (O1) [draw, circle, below=0.55cm of O] {$O_i$};
            \node (Yj) [draw, circle, below=3.2cm of X, xshift=5.8cm] {$C''_i$};
            \node (Source) [draw, ellipse, below=1.2cm of Yj] {$Source_i$};
            \node (help) [below=0.1cm of R, xshift=-0.8cm] {};
            \node (Yjprime) [draw, circle, above=3.1cm of Yj] {$C'_i$};
            \node (Yjreal) [draw, circle, right=1.2cm of Yj] {$C_i$};
    
            \node[draw, thick, fit=(C)(R)(help)(O)(Rprime)(Pij), inner sep=0.3cm, label=below:{$j \in \{1..n_C\} \setminus \{i\}$}] {};
            \node[draw, thick, fit=(R)(Rprime), inner sep=0.1cm, label=above:{$k \in \{1..n_R\}$}] {};
    
            \draw[draw, thick] (X) -- (E);
            \draw[draw, thick] (Rprime) -- (R);
            \draw[draw, thick] (O) -- (R);
            \draw[draw, thick] (O1) -- (R);
            \draw[draw=brown, thick] (E) -- (Sj);
            \draw[draw=red, thick] (E) -- (Yjprime);
            \draw[draw, thick] (Yjprime) -- (Yjreal);
            \draw[draw, thick] (Yj) -- (Yjreal);
            \draw[draw=blue, thick] (C) -- (Yj);
            \draw[draw=brown, thick] (Sj) -- (Yj);
            \draw[draw, thick] (R) -- (Pij);
            \draw[draw=blue, thick] (R) -- (Yj);
            \draw[draw=black, thick] (R) -- (Source);
            \draw[draw, thick] (Source) -- (Yjreal);
            \draw[draw=brown, thick] (C) -- (Sj);
            \draw[draw=brown, thick] (Pij) -- (Sj);
            \end{scope}
        \end{tikzpicture}
    \end{subfigure}
    \caption{Part of the probabilistic graphical model for computing a single node $C_i$ during training, extended with additional variables. $E$ is an embedding represented by a delta distribution. $Parent_{ij}$ is a Bernoulli denoting whether $j$ is a parent of $i$. $C'_i$ denotes prediction of the concept by the encoder, while $C''_i$ is prediction using the rules in the memory. $Source_i$ is a Bernoulli denoting whether $i$ is a source concept, and serves as a "selection" between $C'_i$ and $C''_i$. Each $O_j$ is a delta distribution enforcing a topological ordering.}
    \label{app_fig:pgm_extended}
\end{figure}
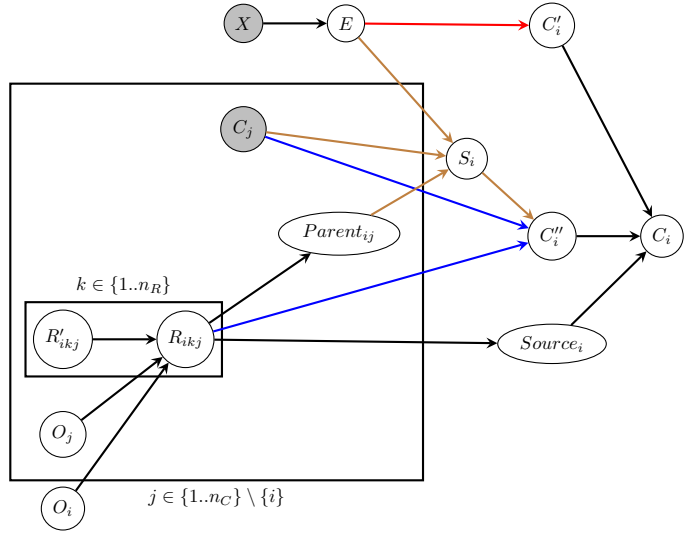

\subsection{General}

In Figure \ref{app_fig:pgm}, we give the probabilistic graphical model for \acr{} during training (Figure \ref{app_fig:pgm_training}), during inference for non-source concepts (Figure \ref{app_fig:pgm_inf_non_source}) and for source concepts (Figure \ref{app_fig:pgm_inf_source}). Additionally, we provide an 'extended PGM' in Figure \ref{app_fig:pgm_extended}, where we add the additional variables that we use in some of the equations (which are effectively marginalized out in Figure \ref{app_fig:pgm}).

\bigskip \noindent
We choose the following conditional probabilities:
\begin{align}
    p(E \mid \hat{x}) &= \delta(E-\hat{e}) \quad \text{where} \quad \hat{e} = g(\hat{x}) \\
    p(\hat{c}_i \mid \hat{c}''_{i}, \hat{c}'_{i}, \hat{source}_i) &= \mathbbm{1}[\hat{source}_i = 1] \cdot \mathbbm{1}[\hat{c}'_i = \hat{c}_i] + \mathbbm{1}[\hat{source}_i = 0] \cdot \mathbbm{1}[\hat{c}''_i = \hat{c}_i] \\
    p(C'_i=1 \mid \hat{e}) &= f_i(\hat{e}) \\
    p(C''_i =1 \mid S_i=k, \hat{c}, \hat{r}_{i,k}) &= l(\hat{c}_{parents(i)}, \hat{r}_{i,k}) \\
    p(S_i=k \mid \hat{x}, \hat{c}, \hat{parent}_{i}) &= h_{i,k}(\hat{x}, \{ \hat{c}_j \mid \hat{parent}_{ij}=1 \}) \\
    p(Source_i = 1 \mid \hat{r}_{i}) &= \prod_{j=1}^{n_C} \prod_{k=1}^{n_R} \mathbbm{1}[\hat{r}_{i,k,j}=I] \\
    p(Parent_{ij} = 1 \mid \hat{r}_{i,:,j}) &= 1 - \prod_{k=1}^{n_R} \mathbbm{1}[\hat{r}_{i,k,j} = I] \\
    p(R_{i,k,j}=P \mid \hat{r}'_{i,k,j}, \hat{o}_i, \hat{o}_j) &= \mathbbm{1}[\hat{o}_j > \hat{o}_i] \cdot \mathbbm{1}[\hat{r}'_{i,k,j} = P] \\
    p(R_{i,k,j}=N \mid \hat{r}'_{i,k,j}, \hat{o}_i, \hat{o}_j) &= \mathbbm{1}[\hat{o}_j > \hat{o}_i] \cdot \mathbbm{1}[\hat{r}'_{i,k,j} = N] \\
    p(R_{i,k,j}=I \mid \hat{r}'_{i,k,j}, \hat{o}_i, \hat{o}_j) &= \mathbbm{1}[\hat{o}_j > \hat{o}_i] \cdot \mathbbm{1}[\hat{r}'_{i,k,j} = I] + \mathbbm{1}[\hat{o}_j \leq \hat{o}_i] \\
    p(O_j) &= \delta(O_j-\hat{o}_j) \quad \text{where} \quad \hat{o}_j \,\, \text{is a learnable parameter}
\end{align}
where $f_i$, $h_{i,k}$ and $g$ are neural networks parametrizing the logits of a Bernoulli, the logits of a categorical, and the point mass of a delta distribution, respectively. The remaining probability is the categorical distribution $p(R'_{i,k,j})$, whose logits are parametrized by a learnable embedding and a neural network mapping this embedding on the logits.

\subsection{Deriving Equation \ref{eq:learning}}

Now we will derive the likelihood formula during training (Equation \ref{eq:learning}). For brevity, we will abbreviate $Source$ as $Src$ and we denote with $\hat{c}$ an assignment to all concepts \textit{except} $C_i$. When summing over $\hat{r}$, we mean summing over all possible assignments to these variables, which are $n_C$ categoricals each with 3 possible values. First we marginalize out $Src$, $C'_i$ and $C''_i$:
\begin{align}
    p(\hat{c}_i \mid \hat{c}, \hat{x}) &= \sum_{\hat{src}_i=0}^{1} \sum_{\hat{c}''_i=0}^{1} \sum_{\hat{c}'_i=0}^{1} p(\hat{c}_i \mid \hat{src}_i, \hat{c}'_i, \hat{c}''_i) \cdot p(\hat{src}_i, \hat{c}'_i, \hat{c}''_i \mid \hat{c}, \hat{x})
\end{align}
We now marginalize $E$ and $R$, exploiting the conditional independencies that follow from the PGM:
\begin{align}
    p(\hat{c}_i \mid \hat{c}, \hat{x}) &= \int_{\hat{e}} \sum_{\hat{r}} \sum_{\hat{src}_i=0}^{1} \sum_{\hat{c}''_i=0}^{1} \sum_{\hat{c}'_i=0}^{1} p(\hat{c}_i \mid \hat{src}_i, \hat{c}'_i, \hat{c}''_i) \cdot p(\hat{src}_i, \hat{c}'_i, \hat{c}''_i \mid \hat{c}, \hat{x}, \hat{e}, \hat{r}) \cdot p(\hat{e}|\hat{x}) \cdot p(\hat{r}) \, \text{d}\hat{e}
\end{align}
Given $C$, $X$, $E$ and $R$, it follows from the PGM that $Src_i$, $C_i'$ and $C_i''$ are conditionally independent, and each of the resulting conditional probabilities can be simplified:
\begin{align}
    p(\hat{c}_i \mid \hat{c}, \hat{x}) = \int_{\hat{e}} \sum_{\hat{r}} \sum_{\hat{src}_i=0}^{1} \sum_{\hat{c}''_i=0}^{1} \sum_{\hat{c}'_i=0}^{1} p(\hat{c}_i \mid \hat{src}_i, \hat{c}'_i, \hat{c}''_i) \cdot p(\hat{src}_i \mid \hat{r}) &\cdot p(\hat{c}'_i \mid \hat{e}) \cdot p(\hat{c}''_i \mid \hat{c}, \hat{e}, \hat{r}) \notag \\
    & \cdot p(\hat{e}|\hat{x}) \cdot p(\hat{r}) \, \text{d}\hat{e}
\end{align}
We exploit the fact that $p(E\mid \hat{x})$ is a delta distribution. After applying the delta distribution's sifting property, we obtain:
\begin{align}
    p(\hat{c}_i \mid \hat{c}, \hat{x}) &= \sum_{\hat{r}} \sum_{\hat{src}_i=0}^{1} \sum_{\hat{c}''_i=0}^{1} \sum_{\hat{c}'_i=0}^{1} p(\hat{c}_i \mid \hat{src}_i, \hat{c}'_i, \hat{c}''_i) \cdot p(\hat{src}_i \mid \hat{r}) \cdot p(\hat{c}'_i \mid \hat{e}) \cdot p(\hat{c}''_i \mid \hat{c}, \hat{e}, \hat{r}) \cdot p(\hat{r})
\end{align}
with $\hat{e} = g(\hat{x})$ the point mass of the distribution. After filling in the conditional probability for $p(C_i \mid \hat{src}_i, \hat{c}_i', \hat{c}_i'')$, we have:
\begin{align}
    p(\hat{c}_i \mid \hat{c}, \hat{x}) = \sum_{\hat{r}} \sum_{\hat{src}_i=0}^{1} \sum_{\hat{c}''_i=0}^{1} \sum_{\hat{c}'_i=0}^{1} & (\mathbbm{1}[\hat{src}_i = 1] \cdot \mathbbm{1}[\hat{c}_i'=\hat{c}_i] +  \mathbbm{1}[\hat{src}_i = 0] \cdot \mathbbm{1}[\hat{c}_i''=\hat{c}_i]) \\ 
    & \cdot p(\hat{c}_i \mid \hat{src}_i, \hat{c}'_i, \hat{c}''_i) \cdot p(\hat{src}_i \mid \hat{r}) \cdot p(\hat{c}'_i \mid \hat{e}) \cdot p(\hat{c}''_i \mid \hat{c}, \hat{e}, \hat{r}) \cdot p(\hat{r})
\end{align}
Most terms become zero due to the indicator functions. After simplifying, we have:
\begin{align}
    p(\hat{c}_i \mid \hat{c}, \hat{x}) &= \sum_{\hat{r}} p(\hat{r}) \cdot \left[ p(Src_i=1 \mid \hat{r}) \cdot p(C'_i = \hat{c}_i \mid \hat{e}) + p(Src_i=0 \mid \hat{r}) \cdot p(C''_i = \hat{c}_i \mid \hat{c}, \hat{e}, \hat{r}) \right] \\
    &= \mathbb{E}_{\hat{r} \sim p(R)} \left[ p(Src_i=1 \mid \hat{r}) \cdot p(C'_i = \hat{c}_i \mid \hat{e}) + p(Src_i=0 \mid \hat{r}) \cdot p(C''_i = \hat{c}_i \mid \hat{c}, \hat{e}, \hat{r}) \right]
\end{align}
which corresponds to Equation \ref{eq:learning}.

\subsection{Deriving Equation \ref{eq:inference_complete}}

\bigskip \noindent
Now we will derive the likelihood formula used during inference (Equation \ref{eq:inference_complete}). 
We begin by marginalizing out $\hat{c}$:
\begin{align}
    p(C_i \mid \hat{x}) &= \sum_{\hat{c}} p(C_i \mid \hat{c}, \hat{x}) \cdot p(\hat{c} \mid \hat{x})
\end{align}
where the sum goes over all possible assignments to all concepts (except $C_i$). As explained in Section \ref{sec:inference}, at inference time, we collapse the distributions over roles $p(R)$ on their most likely values $\hat{r}$, which means that each role $R_{i,k,j}$ is observed and fixed. After filling in Equation \ref{eq:learning}:
\begin{align}
    p(C_i \mid \hat{x}) &= \sum_{\hat{c}} (\mathbbm{1}[\hat{src}_i=1] \cdot p(C'_i \mid \hat{e}) + \mathbbm{1}[\hat{src}_i=0]) \cdot p(C''_i \mid \hat{e}, \hat{c}, \hat{r})) \cdot p(\hat{c} \mid \hat{x})
\end{align}
where $\hat{src}_i$ can be deterministically computed from $\hat{r}$. First, we use that $p(C'_i \mid \hat{e})$ is equivalent to $p(C'_i \mid \hat{x})$ as $E$ is a deterministic function of $X$:
\begin{align}
    p(C_i \mid \hat{x}) &= \sum_{\hat{c}} (\mathbbm{1}[\hat{src}_i=1] \cdot p(C'_i \mid \hat{x}) + \mathbbm{1}[\hat{src}_i=0]) \cdot p(C''_i \mid \hat{e}, \hat{c}, \hat{r})) \cdot p(\hat{c} \mid \hat{x})
\end{align}
Then, we remark that the first term is independent of $\hat{c}$, and the second term is only dependent on $\hat{c}_{parents(i)}$, as follows from the conditional probabilities $p(C''_i \mid S_i=k, \hat{c}, \hat{r}_{i,k})$ and $p(S_i \mid \hat{x}, \hat{c}, \hat{parent}_{i})$. Then, we can rewrite the above equation as:
\begin{align}
    p(C_i|\hat{x}) = \sum_{\hat{c}_{parents(i)}} (& \mathbbm{1}[\hat{src}_i=1] \cdot p(C'_i \mid \hat{x}) \notag \\ + & \mathbbm{1}[\hat{src}_i=0] \cdot p(C_i'' \mid \hat{x}, \hat{c}_{parents(i)}, \hat{r}_i)) \cdot p(\hat{c}_{parents(i)} \mid \hat{x})
\end{align}
which corresponds to Equation \ref{eq:inference_complete}.

\section{Additional results}  \label{app:more_results}

\subsection{Learned rules}

\textbf{\acr{} learns meaningful rules (Table \ref{tab:example_rules}).} For instance, the listed rules for CUB show that \acr{} has learned that certain concepts are mutually exclusive (e.g.\ 'black throat' and 'yellow throat' for birds). In MNIST-Addition, \acr{} has also learned that a single MNIST image only contains 1 digit, and the rules of addition are also recognizable.

\begin{table}
\small
\centering
\caption{Example learned rules for CUB, MNIST-Addition and CIFAR10.}
\begin{tabularx}{\textwidth}{ll}
\toprule
\textbf{Example Rule} & \textbf{Intuition} \\
\midrule
$\text{black throat} \leftarrow \neg \text{yellow throat}$ & Mutually exclusive concepts \\
$\text{black throat} \leftarrow \text{black upperparts} \land \neg \text{yellow throat} \land \text{brown forehead}$ & E.g.\ crows, ravens, blackbirds \\
\midrule
$\text{digit}_1 \, \text{is} \, \text{0} \leftarrow \neg \text{digit}_1 \, \text{is} \, \text{1} \land \neg \text{digit}_1 \, \text{is} \, \text{2} \land ...$ & Mutually exclusive concepts \\
$\text{sum} \, \text{is} \, \text{7} \leftarrow \text{digit}_1 \, \text{is} \, \text{5} \land \text{digit}_2 \, \text{is} \, \text{2} \land \neg \text{digit}_1 \, \text{is} \, \text{1} \, \land ...$ & $5+2=7$ \\
\midrule
$\text{a dock} \leftarrow \text{a port} \land \text{a tire}$ & Ports have docks \\
$\text{a passenger} \leftarrow \text{a cab for the driver}$ & Cabs and their passengers \\
$\text{engines} \leftarrow \text{four wheels}$ & Cars \\
$\text{hooves} \leftarrow \text{long neck}$ & Horses \\
\bottomrule
\end{tabularx}
\label{tab:example_rules}
\end{table}

\subsection{Different intervention policies}

In this ablation study, we investigate \acr{}'s intervenability when using different intervention policies. We consider three policies:

\begin{itemize}
    \item \textbf{Graph-based policy:} this policy is based on \acr{}'s learned concept graph, first intervening on the source concepts and gradually moving to the sinks. This is a natural choice as intervening on earlier concepts in the graph may have a larger impact. This approach is also used by \citet{dominici2024causal} for CGMs, and makes it intuitively easy to interpret results, as the intervention order is the same for different models (i.e.\ if we intervene on 3 concepts, they are the same concepts for \acr{} and all competitors). To ensure the comparison with CGMs is fair, we make sure that the CGMs learn using the same graph that \acr{} learned.
    \item \textbf{Uncertainty-based policy:} this policy uses the uncertainty of the concept predictions, first intervening on concepts with high uncertainty. This policy is introduced by \citet{vandenhirtz2024stochastic} for SCBMs. Note that the intervention order differs between different input examples, and differs between different models.
    \item \textbf{Random policy:} this policy randomly generates an intervention order.
\end{itemize}

\begin{figure}
    \centering
    \includegraphics[width=.99\linewidth]{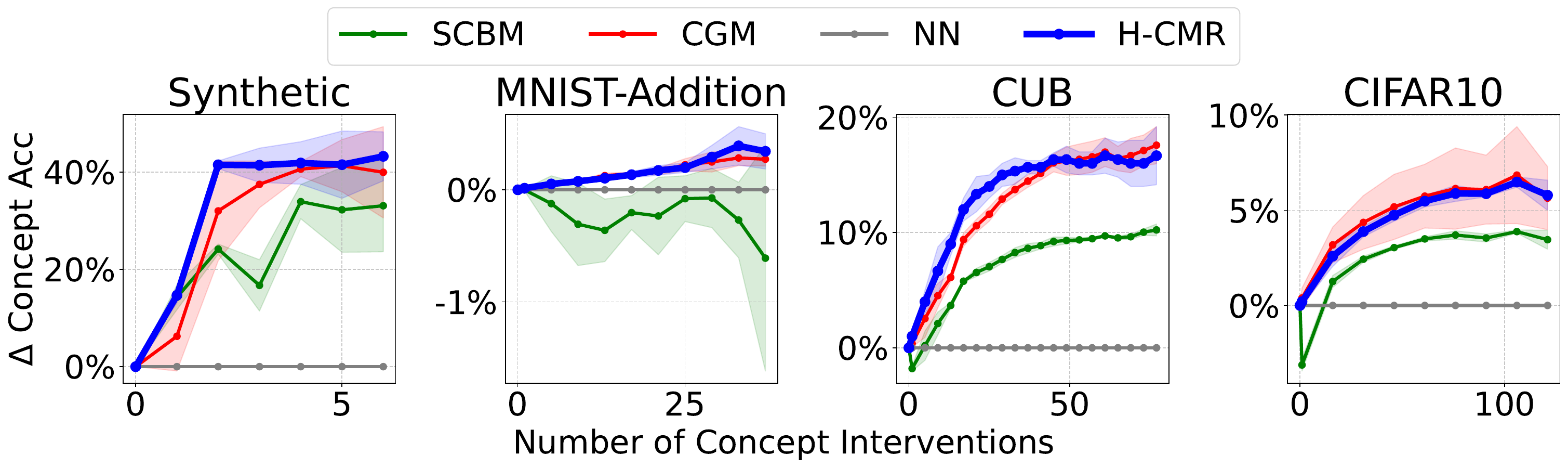}    
    \caption{Difference in accuracy on non-intervened concepts using the graph-based policy after intervening on increasingly more concepts.}
    \label{fig:interventions_graph_delta}
\end{figure}

\bigskip \noindent
We report concept accuracy after different numbers of interventions (like in the main text). For the results in the paper, we additionally report the difference in accuracy after versus before intervening, measured on \textit{non-intervened concepts}, after different numbers of interventions (Figure \ref{fig:interventions_graph_delta}). The former is reported by e.g.\ \citet{vandenhirtz2024stochastic} for SCBM, while the latter is reported by e.g.\ \citet{dominici2024causal} for CGM. 

\bigskip \noindent
Note that in the main text, we report concept accuracy using the graph-based policy.

\begin{figure}
    \centering
    \includegraphics[width=.99\linewidth]{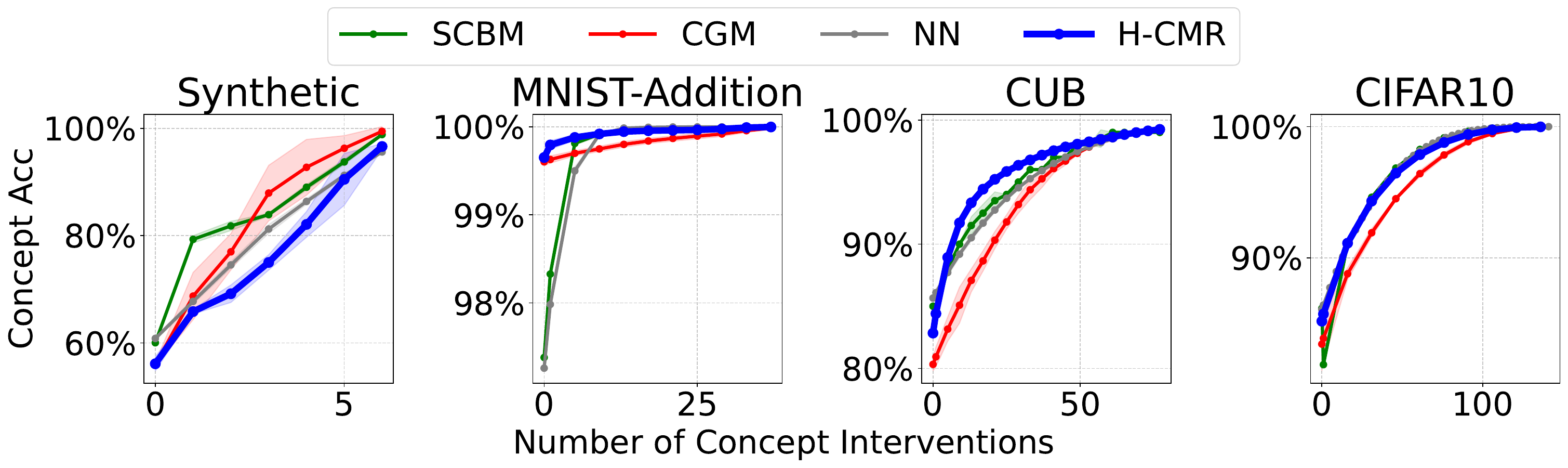}    
    \caption{Concept accuracy using the uncertainty-based policy after intervening on increasingly more concepts.}
    \label{fig:interventions_uncertainty}
\end{figure}

\begin{figure}
    \centering
    \includegraphics[width=.99\linewidth]{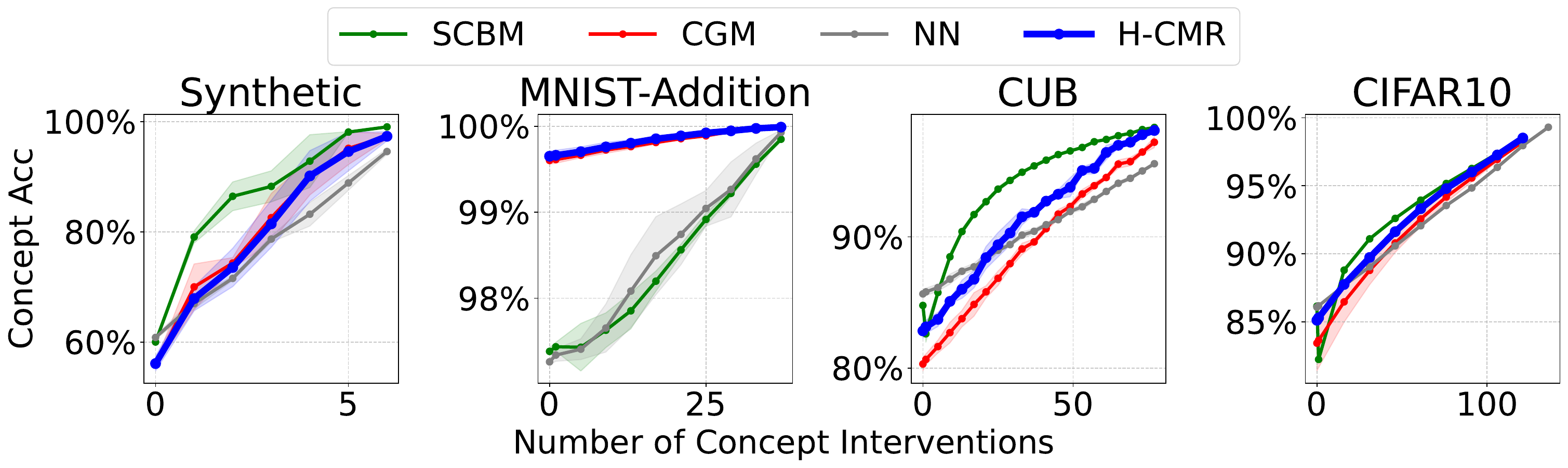}    
    \caption{Concept accuracy using the random policy after intervening on increasingly more concepts.}
    \label{fig:interventions_random}
\end{figure}

\bigskip \noindent
\textbf{\acr{} performs well using other intervention policies (Figures \ref{fig:interventions_uncertainty} and \ref{fig:interventions_random}).}  Using SCBM's uncertainty-based policy (Figure \ref{fig:interventions_uncertainty}), \acr{} performs better than competitors except for the synthetic dataset.  When using a random policy (Figure \ref{fig:interventions_random}), \acr{} performs similar to CGM. For instance, on CUB, \acr{} outperforms NN and CGM but is outperformed by SCBM. On other datasets except the synthetic one, \acr{} performs similar to SCBM and CGM. Note that, as SCBM does not predict concept in a hierarchical fashion, unlike CGM and \acr{}, SCBM is not inherently at a disadvantage when using a random policy.

\section{Proofs}  \label{app:proofs}

\subsection{Theorem \ref{th:expressivity}}

We prove that \acr{} is a universal binary classifier if $n_R \geq 2$ in a similar fashion as done by \citet{debot2024interpretable} for CMR.

\begin{proof}
Source concepts are directly predicted from the input $X$ using a neural network, which is a universal binary classifier. For non-source concepts and tasks $C_i$, the prediction is made by using a neural network to select a logic rule based on the parent concepts $C_{parents(i)}$ and the input $X$, which is then evaluated on the parent concepts $C_{parents(i)}$. Let $C_{k}$ be an arbitrarily chosen parent of $C_i$. Consider the following two rules, which can be expressed in \acr{} as shown between parentheses:
\begin{align}
    C_i &\leftarrow C_{k} \,\,\,\,\, \text{(i.e.\ $\hat{r}_{i,0,k}=P$, \,\, $\forall j \not = k: \hat{r}_{i,0,j}=I$)} \\
    C_i &\leftarrow \neg C_{k} \,\,\,\,\, \text{(i.e.\ $\hat{r}_{i,1,k}=N$, \,\, $\forall j \not = k: \hat{r}_{i,1,j}=I$)}
\end{align}
By selecting one of these two rules, the rule selector neural network can always make a desired prediction for $C_i$ based on the predicted parent concepts $C_{parents(i)}$ and the input $X$. To predict $C_i=1$, the first rule can be selected if $C_k$ is predicted \textit{True}, and the second rule if it is predicted \textit{False}. To predict $C_i=0$, the opposite rule can be selected: the first rule if $C_k$ is predicted \textit{False}, and the second rule if it is predicted \textit{True}.
\end{proof}

\subsection{Theorem \ref{th:expressivity_graph}}

For proving this theorem, we have to prove two statements: that \acr{}'s parametrization can represent any DAG, and that \acr{}'s parametrization guarantees that the directed graph implied by the rules is acylic.

\bigskip \noindent
\textbf{Any DAG over the concepts can be represented by \acr{}:}
\begin{align} 
\forall \, G \in \mathcal{DAG}, \exists \, \theta \in \Theta: \mathcal{G}_\theta=G 
\end{align}
\begin{proof} We will show that for any graph $G$ over the concepts, there exists a set of parameters for \acr{} such that \acr{} has that graph. This means we need to prove that any possible set of edges that form a DAG, can be represented by \acr{}'s parametrization.

\bigskip \noindent
Let $E_G$ be $G$'s set of edges.
For any $G$, it is known that we can find a topological ordering $T_G$ of $G$'s nodes such that for every edge $(i,j) \in E_G$, $i$ occurs earlier in the ordering than $j$. Let $index(i, T_G)$ denote the index of $i$ in this ordering.

\bigskip \noindent
The following random variables of \acr{} are parameters that define the concept graph: $O \in \mathbb{R}^{n_C}$ (delta distributions, parametrized by an embedding), $R'_{i,k,j} \in \{P,N,I\}$ (categorical variables, parametrized by embeddings and neural networks).

\bigskip \noindent
The following assignment to these variables ensures that \acr{}'s concept graph corresponds to $G$:
\begin{align}
    \forall i \in & [1, n_C] : O_i = index(i, T_G)  \label{eq:proof1_3} \\
    \forall i,j \in & [1, n_C], k \in [1, n_R] : p(R'_{i,k,j}=P) =  
        \begin{cases}
            1 & \text{if $(i,j) \in E$} \\
            0 & \text{if $(i,j) \notin E$}
        \end{cases}  \label{eq:proof1_2} \\
    \forall i,j \in & [1, n_C], k \in [1, n_R] : p(R'_{i,k,j}=N) = 0 \\
    \forall i,j \in & [1, n_C], k \in [1, n_R] : p(R'_{i,k,j}=I) = 1 - p(R'_{i,k,j}=P)
\end{align}

\bigskip \noindent
As the parent relation defines the edges in the graph, we need to prove that:
\begin{align}  \label{eq:proof1_1}
    \forall i,j \in & [1, n_C]: Parent_{ij}=1 \Leftrightarrow (i,j) \in E
\end{align}
From \acr{}'s parametrization, we know that:
\[ Parent_{ij}=1-\prod_{k=1}^{n_R} \mathbbm{1}[R_{i,k,j}=I] \]
Thus, it follows that:
\begin{align}
    Parent_{ij}=1 & \Leftrightarrow \exists k \in [1, n_R]: R_{i,k,j} \neq I  \\
                  & \Leftrightarrow \exists k \in [1, n_R]: I \neq \underset{r \in \{P, N, I\}}{\text{arg max}} \, p(R_{i,k,j}=r)
\end{align}
where we used Equation \ref{eq:roles_argmaxed}.
After plugging this into Equation~\ref{eq:proof1_1}, we still need to prove that:
\begin{align}  \label{eq:proof1_5}
    \forall i,j \in & [1, n_C]: (\exists k \in [1, n_R]: I \neq \underset{r \in \{P, N, I\}}{\text{arg max}} \, p(R_{i,k,j}=r)) \Leftrightarrow (i,j) \in E
\end{align}

\bigskip \noindent
We first prove the $\Leftarrow$ direction. If $(i,j) \in E$, then $i$ must appear before $j$ in $T_G$. Therefore:
\begin{align}  \label{eq:proof1_4}
    index(i,T_G) < index(j,T_G)
\end{align}
If we take Equation \ref{eq:true_roles} and fill in Equation \ref{eq:proof1_2}, we know that for all $k$: 
\begin{align}
    p(R_{i,k,j}=P) &= \mathbbm{1}[O_j>O_i] \cdot p(R'_{i,k,j}=P) = \mathbbm{1}[O_j>O_i] \\
    p(R_{i,k,j}=N) &= \mathbbm{1}[O_j>O_i] \cdot p(R'_{i,k,j}=N) = 0 \\
    p(R_{i,k,j}=I) &= \mathbbm{1}[O_j>O_i] \cdot p(R'_{i,k,j}=I) + \mathbbm{1}[O_j \leq O_i] = \mathbbm{1}[O_j \leq O_i]
\end{align}
Then, filling in Equation \ref{eq:proof1_3} and using Equation \ref{eq:proof1_4}, we get
\begin{align}
    p(R_{i,k,j}=P) &= \mathbbm{1}[index(j,T_G) > index(i,T_G)] = 1 \\
    p(R_{i,k,j}=N) &= 0 \\
    p(R_{i,k,j}=I) &= \mathbbm{1}[index(j,T_G \leq index(i,T_G] = 0
\end{align}
which proves the $\Leftarrow$ direction after applying this to Equation \ref{eq:proof1_5}. We now prove the $\Rightarrow$ direction by proving that 
\begin{align}
\forall i,j \in [1, n_C]: (i,j) \not \in E \Rightarrow (\forall k \in [1, n_R]: I = \underset{r \in \{P, N, I\}}{\text{arg max}} \, p(R_{i,k,j}=r)). 
\end{align}
Using Equation \ref{eq:true_roles} and filling in Equation \ref{eq:proof1_3}, we know that for all $k$:
\begin{align}
    p(R_{i,k,j}=P) &= \mathbbm{1}[O_j>O_i] \cdot p(R'_{i,k,j}=P) = 0 \\
    p(R_{i,k,j}=N) &= \mathbbm{1}[O_j>O_i] \cdot p(R'_{i,k,j}=N) = 0 \\
    p(R_{i,k,j}=I) &= \mathbbm{1}[O_j>O_i] \cdot p(R'_{i,k,j}=I) + \mathbbm{1}[O_j \leq O_i] = \mathbbm{1}[O_j>O_i] + \mathbbm{1}[O_j \leq O_i] = 1
\end{align}
which proves the $\Rightarrow$ direction.
\end{proof}

\bigskip \noindent
\textbf{In \acr{}, the directed graph implied by the rules is always acylic:}
\begin{align} 
\forall \theta \in \Theta: \mathcal{G}_\theta \, \, \text{is a DAG}
\end{align}
\begin{proof}
A graph $G$ is a DAG if and only if there exists a topological ordering $T_G$ for that graph, such that for every edge $(i,j) \in E_G$, $i$ occurs earlier in $T_G$ than $j$, with $E_G$ the set of edges of $G$. Thus, we must simply prove that there exists such a topological ordering for each possible $\theta \in \Theta$. We will prove that the node priority vector $O$ defines such an ordering.

\bigskip \noindent
Specifically, let $T_O := \text{arg sort}(O)$ be the concepts sorted based on their value in $O$. We will prove that $T_O$ forms the topological ordering for $\mathcal{G}$.

\bigskip \noindent
We know that
\begin{align}
    T_O \, \text{is a topological ordering for} \, \mathcal{G} \Leftrightarrow \forall (i,j) \in E_{\mathcal{G}}: O_i < O_j
\end{align}
As the parent relation defines the edges in the graph, we need to prove that:
\begin{align}
    \forall i,j \in [1, n_C]: Parent_{ij}=1 \Rightarrow O_i < O_j
\end{align}
for which we will use a proof by contradiction. Let us assume that the above statement is false. Then, we know that:
\begin{align}
    \exists i,j \in [1, n_C]: Parent_{ij}=1 \land  O_i \geq O_j
\end{align}
From \acr{}'s parametrization, we know that:
\[ Parent_{ij}=1-\prod_{k=1}^{n_R} \mathbbm{1}[R_{i,j,k}=I] \]
Thus, we know that:
\begin{align}
    \exists i,j \in [1, n_C], k \in [1, n_R]: R_{i,j,k} \neq I \land  O_i \geq O_j
\end{align}
Then, using Equation \ref{eq:roles_argmaxed}, this is equivalent to:
\begin{align}
    \exists i,j \in [1,n_C], k \in [1, n_R]: I \neq \underset{r \in \{P, N, I\}}{\text{arg max}} \, p(R_{i,k,j}=r)) \land  O_i \geq O_j
\end{align}
Filling in $O_i \geq O_j$ in Equation \ref{eq:true_roles} gives: 
\begin{align}
    p(R_{i,k,j}=P) &= \mathbbm{1}[O_j>O_i] \cdot p(R'_{i,k,j}=P) = 0 \\
    p(R_{i,k,j}=N) &= \mathbbm{1}[O_j>O_i] \cdot p(R'_{i,k,j}=N) = 0 \\
    p(R_{i,k,j}=I) &= \mathbbm{1}[O_j>O_i] \cdot p(R'_{i,k,j}=I) + \mathbbm{1}[O_j \leq O_i] = 1
\end{align}
which is a contradiction, as $I$ is the most likely role.
\end{proof}

\section{Code, licenses and resources}

Our code will be made publicly available upon acceptance under the Apache license, Version 2.0. We implemented \acr{} in Python 3.10.12 and additionally used the following libraries: PyTorch v2.5.1 (BSD license) \cite{paszke2019pytorch}, PyTorch-Lightning v2.5.0 (Apache license 2.0), scikit-learn v1.5.2 (BSD license) \cite{scikit-learn}, PyC v0.0.11 (Apache license 2.0). We used CUDA v12.7 and plots were made using Matplotlib (BSD license). 

\bigskip \noindent
We used the implementation of Stochastic Concept Bottleneck Models (Apache license 2.0)\footnote{\url{https://github.com/mvandenhi/SCBM}} and Causal Concept Graph Models (MIT license)\footnote{\url{https://github.com/gabriele-dominici/CausalCGM}}.

\bigskip \noindent
The used datasets are available on the web with the following licenses: CUB (MIT license\footnote{\url{https://huggingface.co/datasets/cassiekang/cub200_dataset}}), MNIST (CC BY-SA 3.0 DEED), CIFAR10 (MIT license)\footnote{\url{https://www.kaggle.com/datasets/ekaakurniawan/the-cifar10-dataset/data}}.

\bigskip \noindent
The experiments were run on a machine with an NVIDIA GeForce RTX 3080 Ti, Intel(R) Xeon(R) Gold 6230R CPU @ 2.10GHz and 256 GB RAM. Table \ref{tab:computation_time} shows the estimated total computation time for a single run of each experiment.

\begin{table}[h]
\centering
\caption{Estimated total computation time for a single run of each experiment.}
\begin{tabular}{l c}
\toprule
\textbf{Experiment} & \textbf{Time (hours)} \\
\midrule
CUB & $12.2$ \\
MNIST-Addition & $3.3$ \\
MNIST-Addition (with tasks)& $1.6$ \\
CIFAR10 & $33.1$ \\
Synth & $0.1$ \\
\bottomrule
\end{tabular}
\label{tab:computation_time}
\end{table}

\end{document}